\documentclass[journal]{IEEEtran}
\usepackage{cite}
\usepackage{amsmath,amssymb,amsfonts}
\usepackage{algorithmic}
\usepackage{graphicx}
\usepackage{textcomp}
\usepackage{xcolor}
\usepackage{soul}
\usepackage{booktabs}
\usepackage{subcaption}
\usepackage{array}
\usepackage{balance}
\usepackage{svg}
\usepackage[ruled,vlined]{algorithm2e}
\usepackage[hidelinks]{hyperref}

\newcommand{\orcid}[1]{\href{https://orcid.org/#1}{\includegraphics[width=10pt]{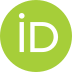}}}

\begin{document}
\title{Spatio-Temporal SAR-Optical Data Fusion for Cloud Removal via a Deep Hierarchical Model}






\author{Alessandro~Sebastianelli \orcid{0000-0002-9252-907X}, Erika~Puglisi, Maria~Pia~Del~Rosso \orcid{0000-0002-0297-0102}, Jamila Mifdal \orcid{0000-0002-9480-7387}, Artur~Nowakowski \orcid{0000-0002-1353-3215}, Fiora Pirri \orcid{0000-0001-8665-9807}, Pierre Philippe Mathieu, Silvia Liberata Ullo \orcid{0000-0001-6294-0581},
\thanks{A. Sebastianelli, M. P. Del Rosso and S. L. Ullo are with the Engineering Department, University of Sannio, Benevento, Italy, email: $\{$sebastianelli, mp.delrosso, ullo$\}$@unisannio.it}
\thanks{A. Nowakowski is with Warsaw University of Technology, Warsaw, masovian, Poland, email: artur.nowakowski@pw.edu.pl}
\thanks{E. Puglisi and F. Pirri are with La Sapienza University, Rome, Italy, email: puglisi.1601231@studenti.uniroma1.it, pirri@diag.uniroma1.it}
\thanks{J. Mifdal and P. P. Mathieu are with the European Space Agency, $\Phi$-lab, Frascati, Italy, email: $\{$jamila.mifdal, pierre.philippe.mathieu$\}$@esa.int}}

\maketitle

\begin{abstract}
Cloud removal is a relevant topic in Remote Sensing as it fosters the usability of high-resolution optical images for Earth monitoring and study. Related techniques have been analyzed for years with a progressively clearer view of the appropriate methods to adopt, from multi-spectral to inpainting methods. Recent applications of deep generative models and sequence-to-sequence-based models have proved their capability to advance the field significantly. Nevertheless, there are still some gaps, mostly related to the amount of cloud coverage, the density and thickness of clouds, and the occurred temporal landscape changes.
In this work, we fill some of these gaps by introducing a novel multi-modal method that uses different sources of information, both spatial and temporal, to restore the whole optical scene of interest. The proposed method introduces an innovative deep model, using the outcomes of both temporal-sequence blending and direct translation from Synthetic Aperture Radar (SAR) to optical images to obtain a pixel-wise restoration of the whole scene. The advantage of our approach is demonstrated across a variety of atmospheric conditions tested on a dataset we have generated and made available. Quantitative and qualitative results prove that the proposed method obtains cloud-free images, preserving scene details without resorting to a huge portion of a clean image and coping with landscape changes.
\end{abstract}
~

\IEEEpeerreviewmaketitle

\section{Introduction}
\begin{figure}[t]
    \centering
    \resizebox{1\columnwidth}{!}{
    \begin{tabular}{cc}
    \includegraphics[width=\columnwidth]{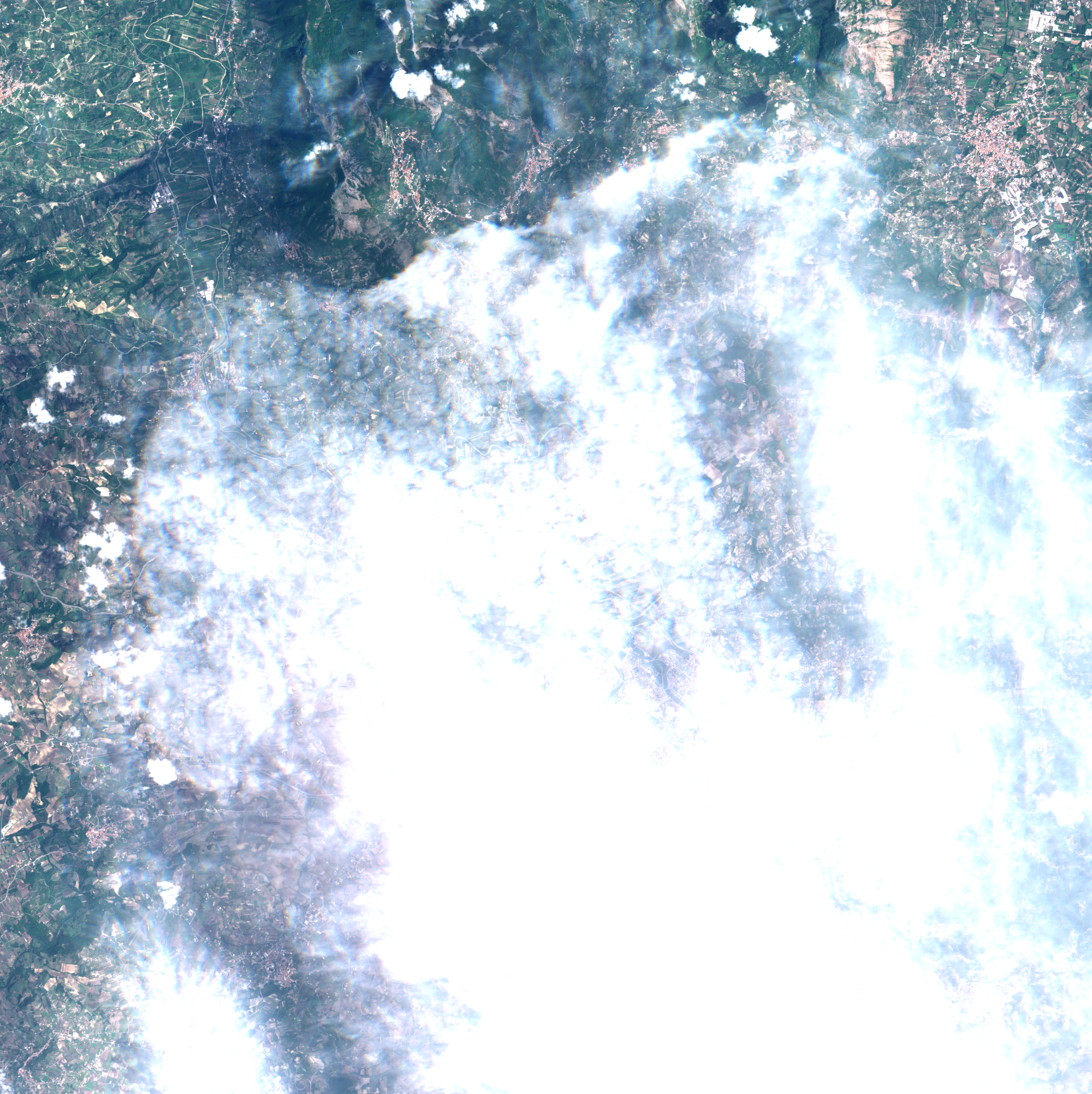} &
    \includegraphics[width=\columnwidth]{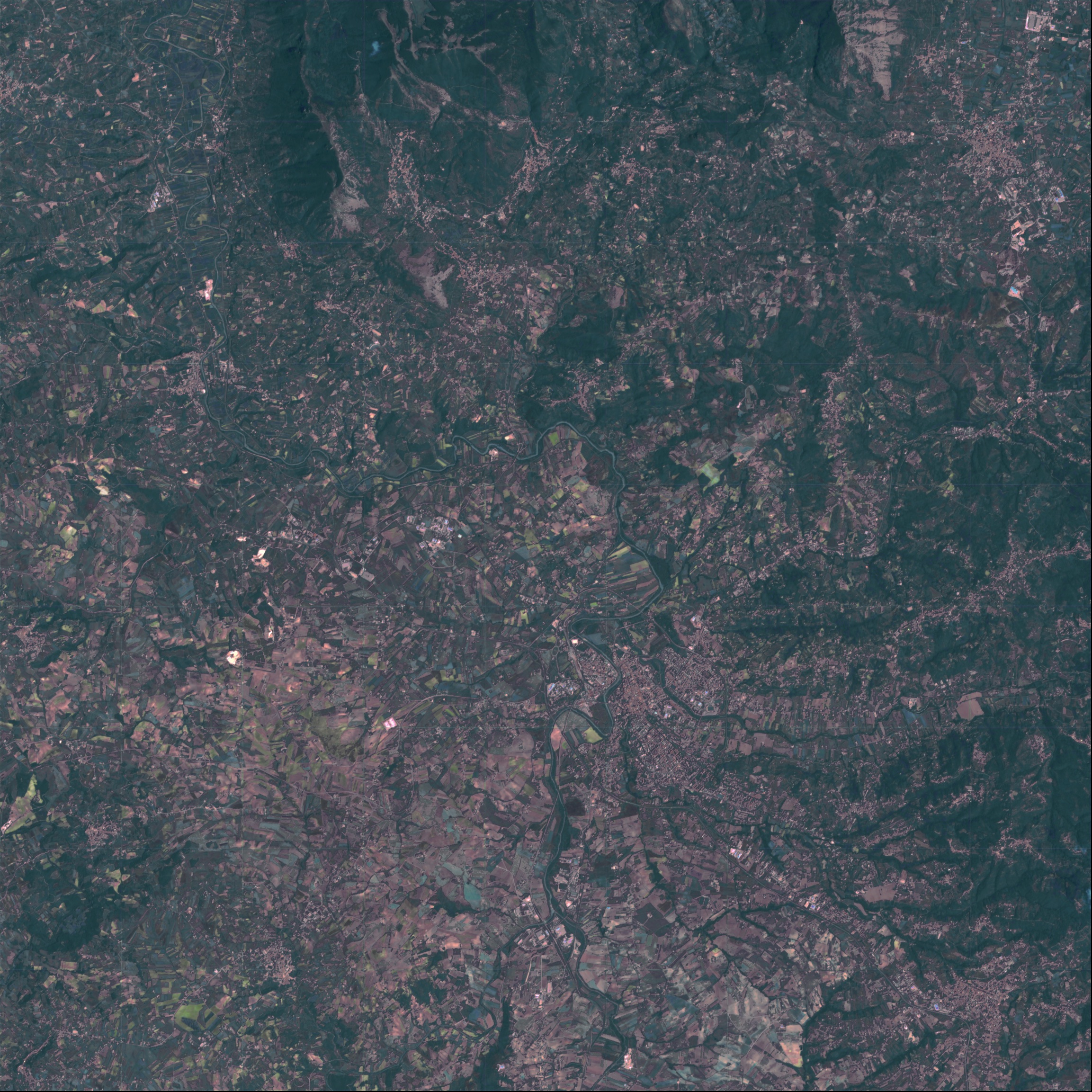}\\
    \scalebox{2.5}{\textbf{(a)}}  & \scalebox{2.5}{\textbf{(b)}}\\ 
    \end{tabular}}
    \caption{Cloud removal with the proposed PLFM model. The Sentinel-2 optical image is completely regenerated without affecting visible pixels, while  keeping the original image high resolution. Cloudy Sentinel-2 image (a) and PLFM prediction (b).}
    \label{fig:results}
\end{figure}

\IEEEPARstart{C}{loud} removal and detection have been studied for a long time with different methodologies, and within different fields, including computer vision \cite{1,2,3,4}, statistics \cite{5}, and multi-disciplinary methods \cite{6,7,8}. Although several remote sensing (RS) 
data acquired by multispectral, hyperspectral or radar sensors are nowadays available for many Earth Observation (EO) use cases, high-resolution (HR) optical images remain the ones mainly employed \cite{prasad2011optical}. Cleaning them from clouds becomes crucial to foster their usability for Earth monitoring.

Recently, deep learning (DL) methods based on convolutional neural networks (CNNs) \cite{9,zhang2018missing,11,12,13}, generative models \cite{14,singh2018cloud,bermudez2018sar,17,18,gao2020cloud,20}, sequence to sequence approaches \cite{21,22,23,24}, and most lately multi-modal methods \cite{25,26,27}, have significantly enhanced the cloud removal techniques, though the problem remains ill-posed. Indeed, there are still several aspects requiring improvements and explanations.

Temporal-sequence-based methods need to account for landscape changes; methods based on pure spatial information still suffer from alignment problems, and reconstruction methods do not yet ensure details preservation without losses in resolution.

On the other hand, metrics are still not entirely adequate. For instance, giving a unique measure for cloud coverage in percentage is misleading as it cannot represent uniform regions and densely varying areas in the same way. Similarly, preserving the details is critical and specific metrics are needed beyond the typical Structural Similarity Index (SSIM) and Peak Signal-to-Noise Ratio (PSNR).

To overcome these drawbacks recent approaches are more oriented towards multimodal methods taking advantages of the availability of different image acquisition resources. Deep learning techniques exploiting multimodality \cite{hong2020more} are now also used in other fields such as, for example, in self-supervised learning \cite{29}. Even if deep multimodal architectures   might seem more complex, they ensure to exploit different sources of information and to capture nonlinear relationships, together with a parsimonious approach to labeling, which is the bottleneck of deep learning.

To overcome many of the drawbacks and limitations just mentioned, in this work we propose a novel multimodal model for cloud removal. In the proposed multimodal model, we introduce Pixel-Level Merging of intermediate feature Maps (PLFM), see Figure \ref{fig:results}, using two sources of information: a temporal sequence of optical images, acquired in a short range to avoid the impact of medium range landscape changes, and a SAR image translated to the corresponding optical one.

The model encodes the transformed images of these two sources of information and reconstructs the cloud-free image with pixel-wise details. The reconstruction process decodes and upsamples the rich features obtained by the temporal and spatial deep methods. We resort to state of the art deep models such as Convolutional Long Short-Term Memory (ConvLSTM) \cite{xingjian2015convolutional} and conditional Generative Adversarial Networks (cGANs) \cite{isola2017image} to build our own model, both introducing specific losses and adapting their architectures to cloud removal task requirements. The proposed method is a multimodal model with three branches maintaining three different loss functions coping with the diversity of input sources. 
However, no image saving is required so that a smooth flow of features from one network to the other is ensured. Several experiments and quantitative and qualitative results demonstrate that the proposed method is competitive with the current state of the art, besides its ability to reconstruct the image of interest under different cloud coverage conditions.

The main contributions of this paper are the following:

\begin{enumerate}
    \item we introduce a multimodal approach to cloud removal. The method defines a multimodal deep model accepting as input both temporal sequences of optical images and  SAR images and predicting a cloud free image. The model requires a limited amount of ground truth images and use a short-range temporal sequence. The model has a three branches architecture and it maintains three different loss functions to cope with the diversity of input sources, though ensuring a flow of features from one branch  to the other.
    \item We provide a novel dataset pairing Sentinel-1 (S1) and Sentinel-2 (S2) images randomly distributed on the whole Earth surface to promote model generalization. The dataset is publicly available \cite{codedataset}. 
\end{enumerate}

The remaining part of the paper is structured as follows. In Sec.~\ref{sec:background} traditional and advanced approaches for cloud removal are presented. In Sec.~\ref{sec:method} we overview the method proposed in this work, presenting in detail the introduced novelties. In Sec.~\ref{sec:experiments} we describe the dataset generation and related aspects. In Sec.~\ref{sec:results_comparisons} we present experiments and results by comparing them with related state-of-the-art results. Finally, in Sec. ~\ref{sec:conclusion_discussion} we discuss limitations and advantages of our approach.  Some mathematical background on state-of-the-art networks referred to in this work is reported in the Appendix \ref{sec:appendix_math_preliminary_GAN_LTM}.

\section{State of the art on Cloud Removal}\label{sec:background}
Cloud removal techniques have been studied for years. Besides traditional ones   Machine Learning (ML)-based methods  are recently advancing this research field.  In the following we discuss both of them. 
\subsection{Traditional approaches for cloud removal}
Traditional approaches can be gathered into three main groups: $1)$ Multi-Spectral, $2)$ Multi-Temporal, and $3)$ Inpainting. 

\noindent
{\bf Multi-Spectral}
 approaches are applied when the optical signals are only partially affected by clouds.   These techniques exploit weak signals to restore the missing surface information without additional data. 
The considered methodologies  are either mathematically grounded \cite{hu2015thin, xu2019thin}, or based on physics \cite{zhu2014continuous, xu2015thin, wang2019detection, lv2016empirical} or on geometry  \cite{choi2004cloud}. 

\noindent
{\bf Multi-Temporal approaches} restore cloudy scenes by integrating information from previous acquisitions with cloud-free conditions, as in \cite{lin2012cloud} \cite{ramoino2017ten, li2015sparse}. Some studies combine data of different types, like for instance in \cite{li2019cloud}, where a cloud-free reference image is obtained by the fusion of two or more low-resolution images coming from different sensors. 

\noindent
{\bf Inpainting approaches} use cloud-detection algorithms to identify the cloudy portion of an image and reconstruct it using the other non-cloudy ones \cite{
Maalouf2009, Cheng2014, meng2017sparse}.
In particular, \cite{siravenha2011evaluating} evaluates two inpainting methods one based on nearest neighbors and the other based on heat-diffusion and splin-plate methods, though they do
not fulfill large damaged regions. 
\subsection{Advanced approaches for cloud removal}
Parallel to the traditional approaches, the availability of a huge amount of data for RS  made it possible to use powerful data-driven methods, such as those based on DL techniques.

\begin{figure*}[!ht]
    \centering
    \includegraphics[width=2\columnwidth]{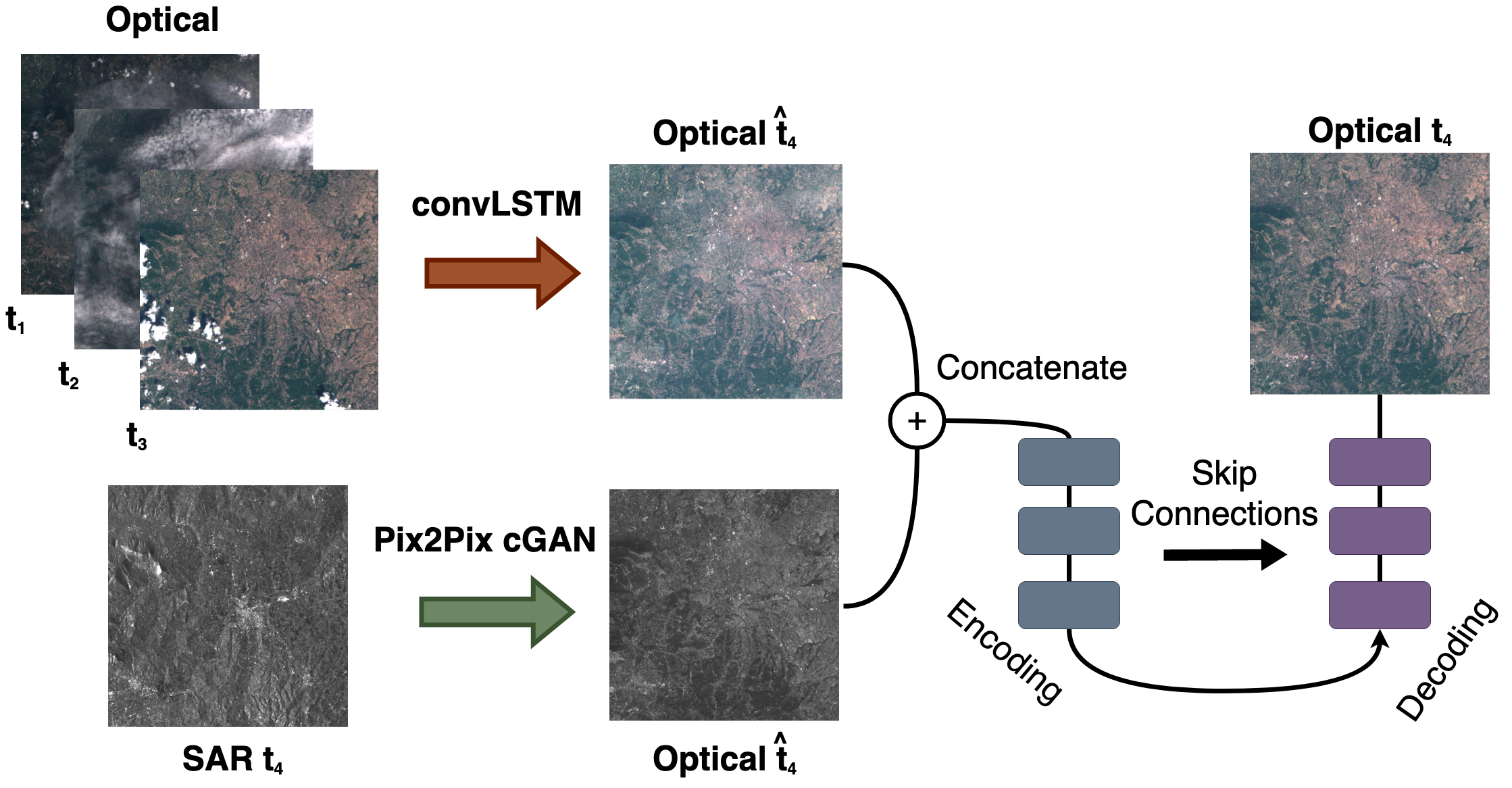}
    \caption{The multimodal PLFM model with three branches: the  ConvLSTM and cGAN based branches  estimate the input to the head branch generating the cloud-free image.}
    \label{fig:model}
\end{figure*}

The first work exploiting the potential of  CNNs \cite{lecun2015deep} for restoring missing information in RS imagery has been presented in \cite{zhang2018missing}. The authors 
adopted a Spatial-Temporal-Spectral (STS) CNN to restore image gaps with multi-source data. However, STS CNNs cannot effectively reconstruct large-scale areas. Moreover they  can only utilize single temporal images. For this reason, in \cite{zhang2020thick} the same authors presented an improved version of their model handling an arbitrary number of temporal images to remove clouds and cloud shadows. Another study based on optical bi-temporal series for the detection and removal of image portions   covered by clouds was presented in \cite{9099032}.

Other works have deployed GANs \cite{goodfellow2014generative}, and their extensions  \cite{isola2017image, Zhu_2017_ICCV, mirza2014conditional}, in particular, this latter  
turning out to be useful for synthetic image generation.  The potential of generative methods for advancing cloud removal techniques has been shown in \cite{ 9211498}.  In  \cite{grohnfeldt2018conditional} the authors   were the first to introduce cGAN to fuse SAR and optical multi-spectral images, and 
cloud/haze-free optical data were generated from images corrupted by clouds. A similar approach has been also proposed in \cite{meraner2020cloud}, which used a Deep Residual Neural Network to remove clouds from multi-spectral S2 images. In this case, a SAR-optical data fusion method exploited the synergistic properties of the two imaging systems for guiding the reconstruction.

A different kind of deep generative model is suggested in \cite{sarukkai2020cloud} with the intent of capturing correlations across multiple images over an area of interest. The proposed model is a Spatio-Temporal Generator Network (STGAN) able to remove clouds by concatenating three previous image acquisitions. In \cite{oehmcke2020creating} the authors proposed a similar approach, but based on a different idea: firstly the feature information is extracted from the previous time steps via a special gating function, then a U-net model is used to obtain the desired output image. A new  type of GAN has been presented in \cite{pan2020cloud}, introducing a spatial attention mechanism inside the cloud removal task, by proposing a Spatial Attention GAN (SpA-GAN). 

The main limitation of all these networks is the requisite of paired datasets available, with both the cloudy images and their cloud-free counterparts (the ground truth) for training. An interesting work, able to overcome this issue has been presented in \cite{singh2018cloud}, where the authors propose a CycleGAN  which learns the map between cloudy and cloud-free images, without the need of a paired dataset or any additional spectral information.

Other approaches remove clouds from images throughout the transformation from SAR to optical images. For instance, in \cite{he2018multi} an optical image is simulated from either a single SAR image or multi-temporal SAR-Optical images using both a CNN and a cGAN. In \cite{gao2020cloud} the SAR data are translated into simulated optical images using a CNN, then a cGAN model fuses the simulated and real optical images to reconstruct the missing parts to obtain a cloud-free output. In \cite{darbaghshahi2020cloud} two GANs are used: the first one transforms SAR images into optical ones and the second one removes clouds using the transformed images of the first model.

However, generative models adapted to the cloud removal task suffer from prediction instabilities that do not allow a simple and reliable use in RS. This is the main reason why they are rarely used alone and are often combined with other types of networks.

\section{Method}\label{sec:method}
\noindent
{\bf Notation}: 
To ease article comprehension, the notation used exposes tensors and matrices represented by uppercase letters with a plain font, column vectors with lowercase letters and a bold font. Superscripts identify the type of satellite data expressed by the matrix or vector. Conversely, subscripts indicate whether the matrix/vector is an input, an output or a ground truth for the model. Estimated values have a hat symbol placed over the variable.

\vskip 0.5\baselineskip
\noindent
{\bf The PLFM Model}:
Removing clouds from an optical image amounts to predict another optical image, yet exploiting features of images from different domains, as already done by other authors such as \cite{he2018multi,gao2020cloud,darbaghshahi2020cloud}. 
The novelty of our work is that in place of removing the clouds we generate a cloud free image. We do so  using intermediate   temporal and spatial features learned by  two networks, forming the two independent input branches of the proposed PLFM model, illustrated in Figure \ref{fig:model}. Our approach is quite different from all other approaches using generative models, because we generate the new image not yet learning discriminative features, but learning to merge  discriminative features, in turn learned by the two input branches of the model:  at the first step we learn  intermediate features, and in the second step we merge them.
The intermediate features are the predicted images of the two networks, acting as feature maps of an intermediate stage of the  reconstruction of the   cloud-free  image.  In fact, the intermediate feature maps are not even saved and passed directly to the network merging them pixel-by-pixel by encoding layers, and finally enhancing the merged image resolution, by decoding.

The PLFM architecture, shown in Figure \ref{fig:model}, is a multimodal deep model with three branches. The two early branches  learn spatio-temporal features from optical sequences on one hand and the idiosyncratic spatial features of SAR images on the other, while the last branch, namely the head, obtains a high-resolution optical image merging the intermediate feature images  at pixel level.

We devised this architecture considering that SAR and multispectral optical data have significantly different domains, affecting the learned feature spaces, which could spoil the image regeneration if trained directly together.

More specifically, we consider as baselines of our model three deep learning paradigms: a ConvLSTM \cite{shi2015convolutional} embedding the convolution operation inside the LSTM, a Pix2Pix cGAN \cite{isola2017image}, and  an encoder-decoder segmentation   network \cite{Ronneberger2015, Zhou2018, Zhou2020}, defining altogether the PLFM model. Each branch of  PLFM  is described in the following.

\vskip 0.5 \baselineskip
\noindent
\textbf{ConvLSTM for optical-to-optical translation}: The ConvLSTM structure, which is the first branch of PLFM,  takes care of the optical-to-optical domain translation. Namely, taking as input a temporal sequence $(t_1,\ldots,t_{n})$  of optical images, of length $n$, it estimates a new image at time $t_{n{+}1}$ in the same optical domain. This transformation encourages inter-images and intra-images features learning, enhanced by the spatio-temporal structure of ConvLSTM. 
Optical  S2 images are tensors $X^{S2} \in \mathbb{R}^{W {\times} H{\times} B}$, where $W$ and $H$ are respectively the width and the height of the image, expressed in the number of pixels, while $B$ represents the number of spectral bands (in this paper only RGB bands are considered).  The ConvLSTM takes as input a time-ordered sequence of S2 images  $\vec{X}^{S2}_{in} \in \mathbb{R}^{n {\times} W {\times}H{\times} B}$, where $n$  indicates the size of the temporal sequence $(t_1,\ldots,t_{n})$, and it predicts  the next S2 acquisition at time $t_{n+1}$. The predicted tensor $\hat{Y}$ is as closest as possible to  S2 acquisition $Y^{S2}_{out} \in \mathbb{R}^{W{ \times} H{\times} B}$, at time $t_{n+1}$. 
The ConvLSTM architecture is formed by stacking ConvLSTM layers alternated with Max Pooling and Batch Normalization layers and, finally a Conv2D layer. With this configuration, the ConvLSTM model learns both temporal (inter-frames) and spatial (intra-frame) features.  

As a loss function we experimented both the Mean Squared Error (MSE) and the Mean Absolute Error (MAE), by identifying  diverse  advantages, in presence of strong or mild differences at pixel level in the sequence.  Consequently, we have  chosen the Huber Loss combining both of them. 

Let the error $e_i(\theta)$ be defined as $(\hat{Y}_{i,\theta} - Y_{out})$, where we dropped the $S2$ superscript, and $\theta$  the network weights estimating $\hat{Y}_i$. Considering a batch of size $k$, the Huber Loss is defined as
\begin{equation}
\begin{array}{ll}
{\mathcal L}_{\delta}(\theta) = \displaystyle{\frac{1}{k} \left( \sum_{e_i(\theta){\leq}\delta} \frac{1}{2}e_i(\theta)^2 {+} 
\sum_{e_i(\theta){>} \delta} \delta|e_i(\theta)| - \frac{\delta}{2} \right)}
\end{array}
\end{equation}

Here $\delta$ is a parameter that is fixed, e.g $\delta{=}1$, then the minimization with respect to the weights amounts to compute:
\begin{equation}
\begin{array}{ll}
\displaystyle{\frac{\partial \mathcal {L}_{\delta}(\theta)}{\partial \theta} = -\sum_{e_i(\theta)\leq\delta}e_i(\theta) + 
\sum_{e_i(\theta)> \delta} \delta \mbox{ sign } e_i(\theta)}
\end{array}
\end{equation}
to update stochastic gradient descent.
The proposed ConvLSTM takes as input a temporal sequence of size $n{=}3$ and the $n{+}1$ element of the sequence as ground-truth and predicts this last. It is trained with the Adam Optimizer, with an early-stopping ensuring to save the best weights, and a learning rate schedule to modulate the optimizer changes, with a starting value of $1e{-}2$.  We have chosen a batch size of 16 temporal sequences $\vec{X}^{S2}_{in}$ of S2 images. 
\vskip 0.6 \baselineskip

\noindent
\textbf{cGAN for SAR-to-optical translation}: Images from the SAR domain are mapped to the optical domain, by translating S1 SAR images into optical ones, resorting to the cGAN  model introduced in \cite{isola2017image}.  Given a dataset of image pairs made by SAR and normalized gray-level optical data, this network predicts an optical image $\hat{Z}^{S2}$. The cGAN input are tensors $X^{S1}\in \mathbb{R}^{W {\times} H {\times} P}$ corresponding to acquisitions in the SAR domain, where $W$ and $H$ are the same parameters described above, and $P$ is the type of polarization (only Vertical-Vertical (VV) is considered in this paper). If any significant temporal (e.g. seasonal evolution) and spatial (e.g. deforestation or new buildings) changes occur in the optical sequence these changes are captured by the SAR sensor,  since the SAR image is immune to clouds; through the cGAN   this information is disclosed and a new optical image accounting for the occurred land changes is generated. 

We recall that the basic idea of image-to-image translation is to learn a parametric translation function that transforms an input image from a source domain into an image of a target domain. The model, as suggested in \cite{isola2017image}, is made  of a patchGAN classifier as discriminator and an encoder-decoder generator with skip connections \cite{Ronneberger2015}. 
 The patchGAn discriminator splits the image predicted by the generator into patches, in general of size $70{\times}70$, and assigns to each pixel a probability of being a real optical image or not. The objective is that the discriminator, trained to classify optical images, finally assigns to each pixel of the image predicted by the generator a probability of 1 that it is an optical image. On the other hand the generator takes as input a SAR image $X$ and estimates an image $\hat{Z}$, that is given to the discriminator.  Therefore the generator has two tasks, minimizing the distance between the true optical image and the image it predicts using the SAR as input, and maximizing the discriminator loss while minimizing its own loss. 
 
 Removing for clarity the superscripts, we are given $G$ indicating the generator, $D$ the discriminator, $U$ the noise, $Z$  the true optical image and ${\hat{Z}}$ the predicted one. The conditional GAN learns the mapping $G:{X,U}\rightarrow \hat{Z}$, minimizing  the loss:

\begin{equation}\label{eq:gen}
\begin{array}{lll}
{\mathcal L}_{cGAN}(G,D) &{=} &{\mathbb E}_{X,\hat{Z}}[\log D(X,\hat{Z})]{+} \\
    &&{\mathbb E}_{X,U}[\log (1{-} D(X,G(X,U)))] \\  
\end{array}
\end{equation}

\noindent
and the $\ell_1$ distance  between true and predicted optical images:
\begin{equation}\label{eq:gmatch}
\begin{array}{l}
{\mathcal L}_{\ell_1}(G) =  {\mathbb E}_{X,Z, U} [\| Z -G(X,U) \|_1] \\
\end{array}
\end{equation}

\noindent
resulting in the final objective function combining both of them:
\begin{equation}
\begin{array}{l}
G^{\star} = \arg\min_G\max_D {\mathcal L}_{cGAN}(G,D) + \lambda {\mathcal L}_{\ell_1}(G)
\end{array}
\end{equation}

The problem with this formulation is speckle noise affecting SAR images \cite{Sebastianelli2022} and
hindering the generator convergence. To remedy this problem we augmented the image set adding simulated SAR images, obtained by corrupting ground truth optical images  with uncorrelated speckle noise simulated according to \cite{abramov2014methods}. Then we split the discriminator into two components,  $D_1$ and $D_2$. 
The task of $D_1$  is to discriminate between the real optical image and the predicted one, obtained from the generator using the simulated SAR images. The task of $D_2$ deals with optical images generated from the real SAR images. Using more than one discriminator has been used already in other works, as for example in \cite{iizuka2017globally}.
The objective function \eqref{eq:gmatch} is thus transformed as follows:

\begin{equation}\label{eq:gen}
 G^{\star} = \arg\min_G \max_{D_1,D_2} \sum_{j=1,2} \gamma_j {\mathcal L}_{cGAN}(G,D_j) + \lambda {\mathcal L}_{\ell_1}(G) 
\end{equation}

\noindent
Here $\gamma_j, j{\in}{1,2}$ weights $D_1$ and $D_2$, with $\sum_j\gamma_j{=}1$. This multi-task approach to cGAN can be extended to more than two tasks by further partitioning the SAR images.

The proposed cGAN is trained with the Adam Optimizer, with a learning rate of $2e{-4}$, both for the generator and the discriminators. We augmented all the available optical images and used a batch size of 32 S1 acquisitions $\Vec{X}^{S1}$. 

\begin{figure*}[!ht]
    \centering
    \includegraphics[width=1.8\columnwidth]{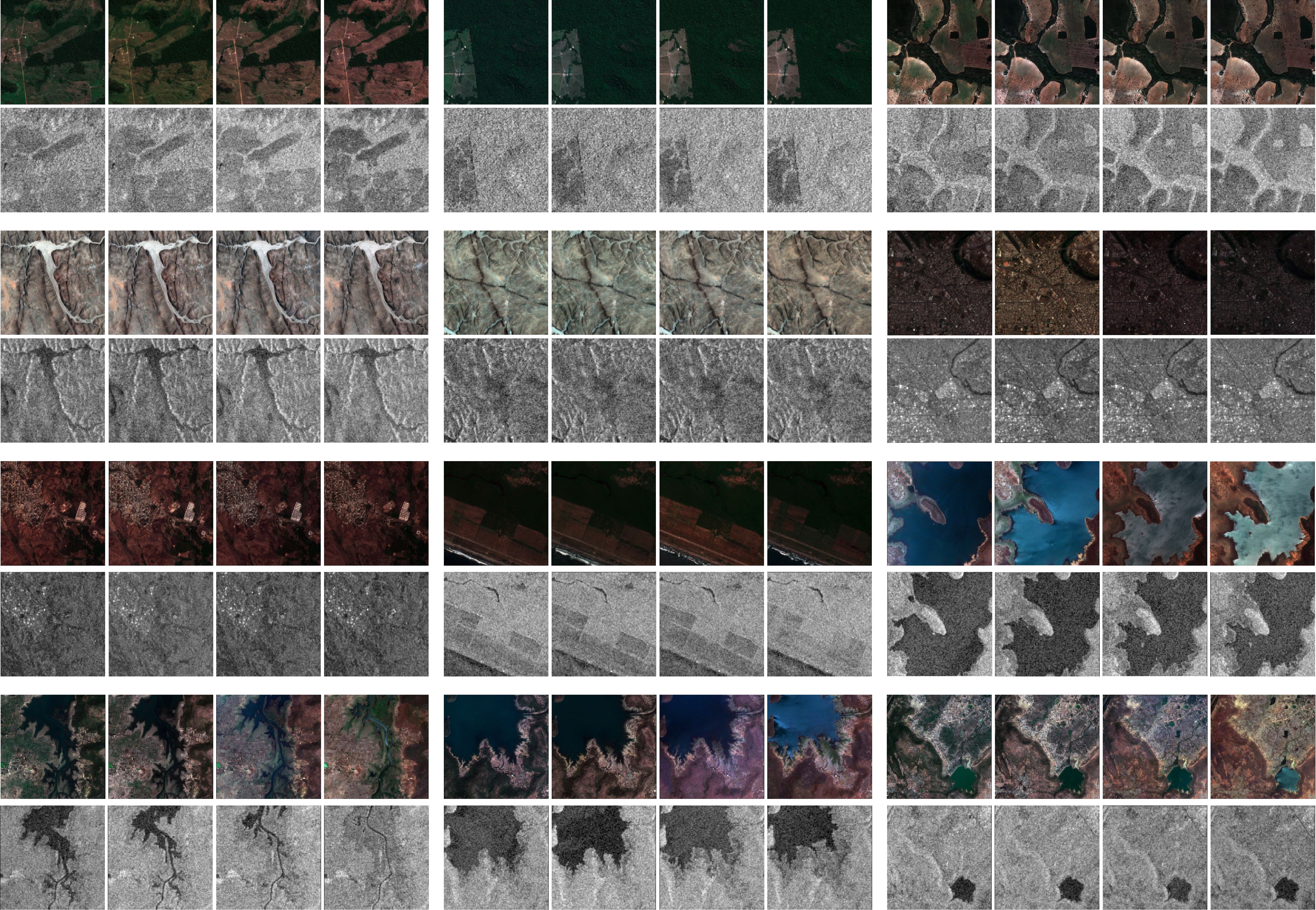}
    \caption{Samples from the dataset for twelve  different geographical regions (randomly selected). For each region time-series of four S2 and corresponding S1 images, acquired with a one month time interval, are shown.}
    \label{fig:s2_s1_dataset_sample}
\end{figure*}


\vskip 0.6 \baselineskip

\noindent
{\bf Head branch of PLFM}:
We consider the outcome of the cGAN and the ConvLSTM to be intermediate feature maps for the head branch of the network  generating a cloud-free image. This last generation step, like semantic segmentation, assigns a class value  to each pixel. We obtain a network that learns to predict, from these intermediate feature maps, an image whose pixels are labeled with the values of a cloud-free image. 

Consider the output ${\hat Y}$ of the convLSTM and the output $\hat{Z}$ of the cGAN. We can consider these outputs as embedding of the inputs $X^{S2}$ and $X^{S1}$. These embeddings are concatenated along the channel dimension to form the map ${\mathcal F}$, namely:
\begin{equation}\label{eq:embedding}
    \hat{Z}\oplus_{2} \hat{Y} {=} {\mathcal F}\in {\mathbb R}^{W{\times}H{\times} d}
\end{equation}
Here, $d{=}6$ corresponds to the number of channels of ${\mathcal F}$. Since the network predicts intensity classes it takes as input a concatenated channel at a time, which we indicate as ${\mathcal F}^k, k{=}1,{\ldots},3$. 
The  network $PLFM{:} {\mathcal F}^k\rightarrow M$,  where  $M{\in} {\mathbb R}^{W{\times}H{\times} |C|}$,  is a tensor providing for each pixel $x$ in ${\mathcal F}^k$ the probability to belong to class $c\in C$, and $C=\{0,{\ldots},255\}$ are all the semantic classes, namely intensity values.

Let $V^j$ be a one-hot encoded vector of length $|C|$ with a $1$ at the reference class location and a $0$ everywhere else.  We denote by $v_c^j$ the $j$-th pixel location at the  $c$-th channel of the reference class label in $C$, and by $p(v_c^j){=}\frac{\exp(v_c^j)}{\sum_{i=0}^{|C|}\exp v_i^j}$  the network softmax output. Namely, the network computes an unnormalized vector of length $|C|$ assigning to each pixel in ${\mathcal F}^k$ a score, and the softmax   normalizes this score in $(0,1)$, such that  $\sum_{c=0}^{|C|}p(v_c)=1$. Note that the softmax activation of the network logits ensures that each pixel belongs to only a single class.  We consider the categorical cross entropy loss, which is computed on the softmax output of the network.   Let $N{=}H{\cdot}W$ be the number of pixels, the categorical cross-entropy loss is:  
\begin{equation}
    \centering
    {\mathcal L}_{CE} = -\sum_{c=0}^{|C|} \sum_{j=1}^N v_c^j \log p(v_c^j)
\end{equation}
\noindent
The minimization of the categorical cross entropy amounts to update the network weights by stochastic gradient descent. The predicted tensor $M$ generates the intensity optical image  $\hat{I}_{\iota}$ with intensity values $\iota{\in} RGB$ as follows:
\begin{equation}
\hat{I}_{\iota} = \arg\max_c\{p(v_c^j) |  v_c^j\in M\}
\end{equation}
Clearly, repeating the computation of $M$ for each combined channel in ${\mathcal F}$ corresponding to $\iota$, we finally obtain the newly generated optical image $\hat{I}$.
For training  this  model a batch of 16 couples of ConvLSTM generated images and cGAN generated images are used as feedforward input for the model.  The network is trained with the Adam optimizer with a learning rate of $2e{-4}$.
 
Note that despite the number of classes is large the learning is favored on one side by the balanced pixels distribution,  and on the other side by the fact that the distribution of intensity values in ${\mathcal F}$ is similar to the ground truth optical image at several locations.  The variety of ground truth samples encourages the categorical cross entropy to approximate the target distribution.\\
 

\section{Dataset and related aspects}\label{sec:experiments}
 
The dataset, implemented  for this application, aims to foster the exploration of cross modal and temporal AI-based applications. Specifically we used it to train our PLFM for the cloud removal task. Yet, this dataset may  represent a useful starting point for other case studies.

The generated dataset contains corresponding S1 and S2 acquisitions randomly distributed on the Earth surface, with 141 different Regions of Interest (ROIs). For each geographical region there is a time-series of four images, acquired with a one month time interval.  The dataset has been built using the \textit{Satellite Data Downloader Tool} proposed in \cite{sebastianelli}. The generated dataset can be found at the link reported in \cite{codedataset}, which the interested reader can refer to for more details. 

As basic information for a better comprehension of the further steps, it is worth to specify that each time-series corresponds to a sequence of four S1 or S2 images, acquired in adjacent months. This temporal constraint has been introduced in order to ease the selection of SAR and optical (cloudy or non-cloudy) images, unfortunately by losing in temporal resolution which is much better than one month for S1 and S2.
We organized the dataset so as to ensure that the SAR images time-series  were available, as the optical time-series. Indeed, also the SAR time-series presents a temporal evolution that can be exploited.\\ 
Images in our dataset have a size of  $256\times256$ pixels, and the whole dataset is formed by $17000$ SAR images and $17000$ Optical Images.


In Figure \ref{fig:s2_s1_dataset_sample} samples from the dataset for twelve different geographical regions are shown. Each sample contains a time-series of four S2 images (on the top) and the corresponding S1 images (on the bottom).


From the figure the diversity of the developed dataset, containing acquisitions over cities, lakes, mountains, coastlines, and other geographical details is evident, and this variety has been introduced to increase the input distribution from which the models learn, to avoid biases from some geographical areas.  

Besides that, since the dataset has been created by random acquisitions over the globe, it has been necessary to develop a strategy for the training-validation split to ensure a balance of the ROIs. This is a very important point and a short description is given for a clear comprehension in the following.

\subsection{Training-validation split}
The splitting method was devised as part of our dataset creation, and in particular since we identified some extreme regions, which could affect the model generalization, this iterative procedure has been introduced to counter this effect. 

The dataset has been divided into training and validation sets by minimizing the dissimilarity between the cumulative histograms computed on several random splits. The algorithm is shown below in the frame named "Algorithm \ref{alg:training_validation_split}". 

For each iteration j a random split, $80\%$ for the training dataset and $20\%$ for the validation dataset, is created. N images from both the datasets are used to compute respectively the training cumulative histogram and the validation cumulative histogram, then the dissimilarity between the two histograms is computed by  equation (\ref{eqn:dissimilarity}) and saved in a vector ${\vec{d}}$. 

\begin{equation}
    d_{j} = \frac{\frac{1}{N}\sum_{i=0}^N |\frac{h_{train}}{\# bins}-\frac{h_{val}}{\# bins}|}{\frac{1}{N}\sum_{i=0}^N \frac{h_{train}}{\# bins}}
    \label{eqn:dissimilarity}
\end{equation}
where $h_{train}$ is the cumulative histogram for the training sub-set, $h_{val}$ is the cumulative histogram for the validation sub-set and $\# bins$ is the number of bins used to calculate the histograms.

By selecting the minimum value of the vector ${\vec{d}}$, the algorithm returns the training and validation split with the lower value of dissimilarity.

\begin{algorithm}[!ht]
\SetAlgoLined
    \For{i $\leftarrow$ 1, number of splits}{
        set$^A_i$, set$^B_i$ $\leftarrow$ random split dataset ($80\%$-$20\%$)\;
        set$^A\_$s $\leftarrow$ select N samples of set$^A_i$\;
        set$^B\_$s $\leftarrow$ select N samples of set$^B_i$\;
        set$^A\_$hist $\leftarrow$ cumulative histogram of set$^A\_$s\;
        set$^B\_$hist $\leftarrow$ cumulative histogram of set$^B\_$s\;
        ${\vec{d}_{i}}$ $\leftarrow$ dissimilarity(set$^A\_$hist, set$^B\_$hist)\;
    }
    best$\_$split $\leftarrow$  $argmin\ {\vec{d}_{i}}$\;
    \Return set$^A_{best\_split}$, set$^B_{best\_split}$\;
 \caption{Training-Validation Splitting procedure}
 \label{alg:training_validation_split}
\end{algorithm}

To split the proposed dataset the number of iterations has been set to 2000, with N equals to 150 and number of bins equal to 20. This procedure ensured the right balance of ROIs in the final training and validation sets (80$\%$-20$\%$). The same procedure is applied to the remaining part of the training set to get the test set (10$\%$).
Finally, our dataset is composed of  $12240$ training SAR images, $12240$ Optical Images,   $3400$ validation SAR images and $3400$ optical images, and as test images $1360$  SAR and $1360$  optical images.

Once the splitting procedure is completed, the training of the model can be accomplished. 

\subsection{Evaluation Metrics}\label{sec:metrics}
In this section we describe the evaluation metrics we use  to measure the performances of the proposed model. Each metric measures a specific property of the estimated data. Additionally to these metrics, we define a co-registration shift compensation mechanism, that slightly modifies the metrics to mitigate the inner co-registration errors of the source data. This last aspect is discussed at the end of this section.

Let $\hat{Y}\in \mathbb{R}^{W\times H \times B}$ be a reference image (i.e. ground truth) and $Y \in \mathbb{R}^{W\times H \times B}$ be the predicted image. The quality measures  are the following:


\begin{enumerate}
\item \textbf{PSNR}: \textit{PSNR} (Peak Signal-to-Noise Ratio) measures the quality of the spatial reconstruction of each hyper-spectral band. \textit{PSNR} is the ratio between the maximum power of the $k^{th}$ band of the reference image and the residual error between the $k^{th}$ bands of the reference and of the estimated images. For the $k^{th}$ band, \textit{PSNR} is computed as follows: 
\begin{equation}
\text{PSNR}(\hat{Y}_{k}, Y_{k})=10\log_{10}{\bigg(\frac{\max{(\hat{Y}_{k})}^2}{\|\hat{Y}_{k}-Y_{k} \|_{2}^{2}/(WH)}\bigg)}
\end{equation}
here $\max{(\hat{Y}}_{k})$ is the maximum pixel value of the $k^{th}$ band for the reference image $\hat{Y}$. The residual error is normalized for each band and thus it is not affected by data values, in so ensuring fair comparison between the bands. The final value of \textit{PSNR} is the average of the \textit{PSNR} values taken at each band. An high value of \textit{PSNR} indicates that the reconstruction is of high quality.

\item \textbf{SSIM}: The Structural Similarity Index Measure (\textit{SSIM}) measures the similarity between two images $\hat{Y}$ and $Y$. The comparison is performed on the basis of luminance, contrast and structure of the images as follows:
\begin{equation}
\text{SSIM}(\hat{{Y}}, {Y})= \frac{(2\mu_{\hat{{Y}}}\mu_{{Y}}+C_1)(2\sigma_{\hat{{Y}}{Y}}+C_2)}{(\mu_{\hat{{Y}}}^2 + \mu_{{Y}}^2+C_1)(\sigma_{\hat{{Y}}}^2+\sigma_{{Y}}^2+C_2)}
\end{equation}
Here $C_1$ and $C_2$ are constants, $\mu_{\hat{{Y}}}$, $\mu_{{Y}}$, $\sigma_{\hat{{Y}}}$ and $\sigma_{\textbf{Y}}$ are respectively the means and the standard deviations of the real and of the generated images, and $\sigma_{\hat{{Y}}{Y}}$ is defined as follows:
\begin{equation}
\sigma_{\hat{{Y}}{Y}}=\frac{1}{N-1}\sum_{j=1}^{N}(\hat{{Y}}_j-\mu_{\hat{{Y}}})({Y}_j-\mu_{{Y}}).
\end{equation}
\textit{SSIM} computes values between 0 and 1, where 1 means "very similar" and 0 "very different".

 
\item \textbf{SAM}: \textit{SAM} (Spectral Angle Mapper) \cite{kruse1993spectral} measures the quality of the spectral 
reconstruction by computing the angle between a target $\hat{\mathbf{y}}_{i}$ and a ground truth $\textbf{y}_{i}$ spectral vectors at each pixel in the reference and in the reconstructed image, where the superscript indicates the $i$-th pixel.   \textit{SAM} is measured as follows: 
\begin{equation}
\text{SAM}=\arccos{\bigg(\frac{\hat{\mathbf{y}}_{i}^{\top} \textbf{y}_{i}}{\|\hat{\mathbf{y}}_{i}\|_{2}\|\textbf{y}_{i}\|_{2}}\bigg)}.
\end{equation}
Here $\|\cdot\|_2$ is the Euclidean norm. 
The values of \textit{SAM} are measured in degrees: the smaller the absolute value of \textit{SAM}, the weaker the spectral distortion and the higher the spectral quality of the fusion (the ideal value is 0). The final \textit{SAM} value is computed by averaging all the \textit{SAM}s of the pixels of the image.
\item \textbf{MSE} and \textbf{RMSE}: \textit{MSE} (Mean-Squared Error) is simply the mean squared error between the original image $\hat{{Y}}$ and its estimation ${Y}$ as follows:
\begin{equation}
\text{MSE}(\hat{{Y}}, {Y})= \frac{\sum_{j=1}^{N}(\hat{{Y}}_j-{Y}_j)^2}{N}.
\end{equation}
\textit{RMSE} (Root-Mean-Square Error) is its squared root and measures the $L^{2}$ error between the original image $\hat{{Y}}$ and its estimation ${Y}$ as follows:
\begin{equation}
\text{RMSE}(\hat{{Y}},{Y})=\frac{\|{Y}-\hat{{Y}}\|_{F}} {\sqrt{N\times H}},
\end{equation}
where $\|\hat{{Y}}\|_{F}=\sqrt{\text{trace}(\hat{{Y}}^{\intercal}\hat{{Y}})}$ is the Frobenius norm of $\hat{{Y}}$. The ideal value for \textit{MSE} and \textit{RMSE} is 0.

\item \textbf{CC}: Cross Correlation (\textit{CC}) is defined as follows: 
\begin{equation} 
\text{CC}(\hat{{Y}},{Y})=\frac{1}{H}\sum_{k=1}^{H}\text{CCS}
(\hat{{Y}}_{k}, Y_{k}),
\end{equation}
where CCS is the cross correlation between two single-banded images, \textbf{A} and \textbf{B}, and it is defined as follows:
\begin{equation}
\text{CCS}({A},{B})=\frac{\sum_{j=1}^{N}({A}_{j}-\mu_{A})({B}_ { j } -\mu_ { B } ) } { \sqrt { \sum_ { 
j=1}^{N}({A}_{j}-\mu_{A})^{2}\sum_{j=1}^{N}({B}_{j}-\mu_{B})^{2} }},
\end{equation}
where $\mu_{A}=\frac{1}{N}\sum_{j=1}^{N}{A}_{j}$ is the mean of ${A}$, $\mu_{B}$ is the mean of ${B}$ and N is the number of the pixels. The ideal value of \textit{CC} is 1.

\begin{figure*}[!ht]
    \centering
    \resizebox{2\columnwidth}{!}{
    \begin{tabular}{cccc}
    \includegraphics[width=0.6\columnwidth]{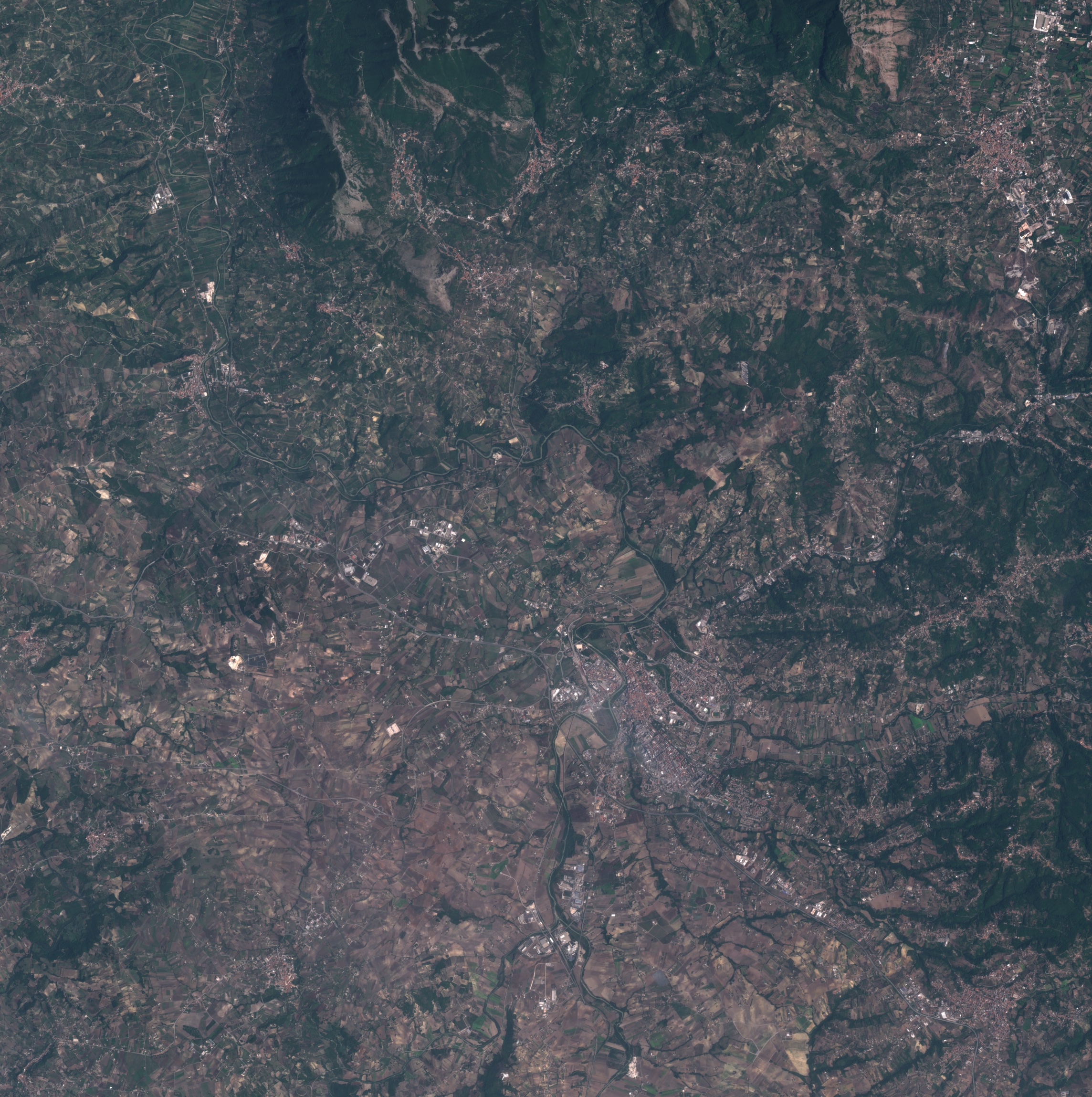} &
    \includegraphics[width=0.6\columnwidth]{imgs/cloudy.png} &
    \includegraphics[width=0.6\columnwidth]{imgs/prediction.jpg}&
    \includegraphics[width=0.6\columnwidth]{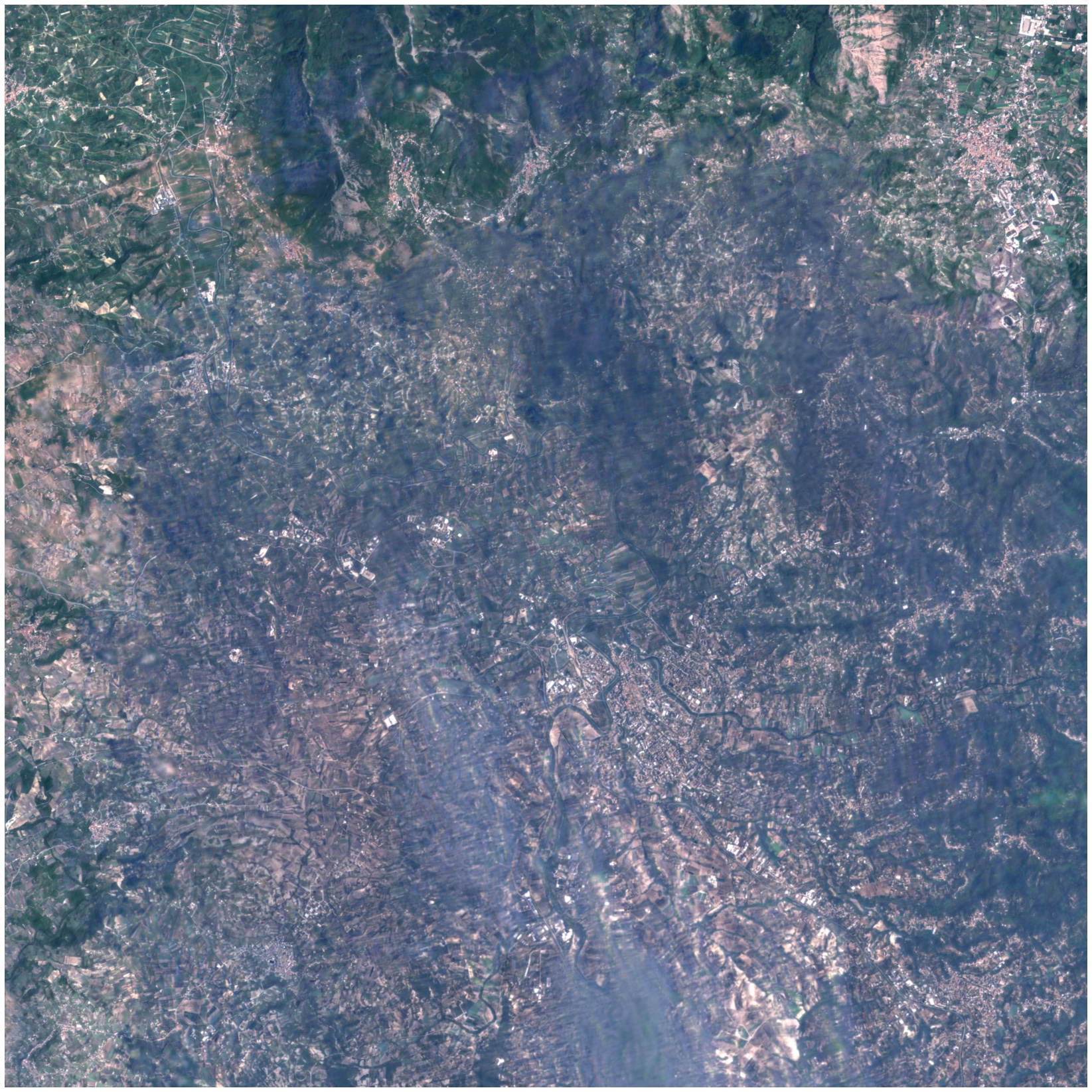}\\
    
    \scalebox{1.2}{\textbf{(a)}} &  \scalebox{1.2}{\textbf{(b)}} & \scalebox{1.2}{\textbf{(c)}} &  \scalebox{1.2}{\textbf{(d)}}\\
    \end{tabular}}
    \caption{Cloud removal with the PLFM model, the Sentinel-2 optical image is completely regenerated. The PLFM model does not affect visible pixels and keep the high resolution of the image. (a) Sentinel-2 ground-truth acquired with a temporal shift, (b) Cloudy Sentinel-2 image, (c) PLFM prediciton and (d) DSen2-CR prediction \cite{meraner2020cloud}.}
    \label{fig:comparison}
\end{figure*}

\item \textbf{DD}: The measure of Degree of Distortion (\textit{DD}) between two images $\hat{{Y}}$ and ${Y}$ is defined as follows:
\begin{equation*}
\text{DD}(\hat{{Y}},{Y})=\frac{1}{N\times H}\|\text{vec}(\hat{{Y}})-\text{vec}({Y})\|_{1},
\end{equation*}
where N is the number of the pixels, H is the number of the bands and $\text{vec}(\hat{{Y}})$ and $\text{vec}({Y})$ represent the vectorization of the images $\hat{{Y}}$ and ${Y}$, respectively. The ideal value of \textit{DD} is 0.

\item \textbf{Q} or \textbf{UQI}: The Universal image Quality Index (\textit{Q} or \textit{UQI}) was suggested by Wang and Bovik~\cite{wang2002universal} to evaluate the similarity between two single-band images. It measures their distortion as the
product of loss of correlation, luminance distortion and contrast 
distortion. The \textit{Q} index between two single-band images ${A}$ and ${B}$ is defined as follows:

\begin{equation}
Q({A},{B})=\frac{4 \sigma_{AB}^{2}\mu_{A}\mu_{B}}{(\sigma_{A}^{2}+\sigma_{B}^{2})(\mu_{A}^{2}+\mu_{B}^{2})}
\end{equation}
where $\mu_{A}, \mu_{B}, \sigma_{A}^{2}$ and $\sigma_{B}^{2}$ are the means and the variances of ${A}$ and ${B}$, respectively, and $\sigma_{AB}$ is the covariance of (${A}$,${B}$). The range of \textit{Q} is $[-1,1]$ and its ideal value is 1. 
\end{enumerate}

\textbf{Co-registration Shift Compensation (CSC)}
As anticipated at the beginning of this section, there may be co-registration errors in the downloaded data, which appear often evident in the time-series. In fact for S2 data, as shown in Table \ref{tab:s2_error}, in 1\% of the cases the co-registration error can be higher than 1.5 pixels \cite{s2_report}. On the other hand, for S1 GRD products, the error should always be below $<0.25$ pixels \cite{s1_report}.

\begin{table}[!ht]
    \centering
    \caption{Multi-temporal performance statistics for S2 constellation \cite{s2_report}.}
    \resizebox{\columnwidth}{!}{
    \begin{tabular}{ccccc}
        
        \toprule
         Co-registration & $0<X<0.5$  & $0.5<X<1$  & $1<X<1.5$  & $X>1.5$ \\
         error &  pixels &  pixels & pixels &  pixels\\
         \midrule
         S2A \% of products &  58\% & 35\% & 6\% & 1\% \\
         S2B \% of products &  47\% & 41\% & 11\% & 1\% \\
         \bottomrule
    \end{tabular}
    }
    \label{tab:s2_error}
\end{table}


The shift compensation is introduced to make the metrics more robust to the co-registration errors and to mitigate their influence during the evaluation of the model. Co-registration errors, producing a misalignment between the pixels of the predicted output and those of the ground truth, affect the models' evaluation.


Nevertheless, coregistrating all the acquisitions would be unfeasible and inefficient. A metric with CSC is calculated by offsetting multiple times the ground truth with respect to the prediction, by one or more pixels. The final value is obtained by maximizing or minimizing the above quantities, depending on how the original metric is defined.

Let $\hat{{Y}}\in \mathbb{R}^{W\times H \times B}$ be a reference image (i.e. ground truth) and let ${Y} \in \mathbb{R}^{W\times H \times B}$ be the predicted image, the generic metric $M$ (e.g. PNSR, MSE, etc.) with the CSC is given by:

\begin{subequations}
    \centering
    \begin{align}
        M1(e_1, e_2) &= M(\hat{{Y}}, {Y_{shifted}(e_1,e_2)})\\
        M_{CSC} &= \min_{e_1,e_2} (or\ \max_{e_1,e_2})\ M1(e_1, e_2)
    \end{align}
\end{subequations}

where ${Y_{shifted}(e_1,e_2)}$ is the shifted version of the predicted image and $e_1,e_2 = {-E,\dots, -1, 0, +1, \dots, E}, E \in N^+$ are the spatial shifts.

\section{Experiments and results}\label{sec:results_comparisons}

Experiments and comparisons are performed on the validation set of the dataset composed of patches with different cloud coverage (e.g. thin, thick and full coverage), ranging from 0 up to $\sim 99\%$ haze/clouds. All the qualitative results shown with the images are done on the test set. 
The intent is to validate and test the proposed method under different conditions. 


In Table \ref{tab:resour} we present the results of the PLFM model, with  several cloud coverage levels (from $20\%$ up to  $100\%$), on the validation set, 
according to the metrics introduced in Section \ref{tab:resour}. We can note that, though increasing the cloud coverage there is a loss of performance  on all the metrics, this is not quite significant. This fact shows that our approach is effectively generalizable to all weather conditions and that it is a promising solution for cloud removal. 

\begin{table*}[!ht]
    \centering
    \caption{Results of PLFM on our val dataset with different cloud conditions}\label{tab:resour}
    \resizebox{1.7\columnwidth}{!}{\begin{tabular}{lccccc}
    \toprule
                 & Cloud Coverage  \\
                 &   $\leq 20\%$ & $20\% < x \leq 50\%$ & $50\% < x \leq 80\%$ & $80\% < x \leq 100\%$\\
    \midrule 
    PSNR $\uparrow$   &  30.239 &  30.061 &  28.092 &  27.146 \\
    SSIM $\uparrow$   &   0.926 &   0.931 &   0.921 &   0.922 \\
    RMSE $\downarrow$ &   0.027 &   0.036 &   0.031 &    0.036 \\
    SAM $\downarrow$  &   0.088 &   0.095 &   0.089 &    0.093 \\
    UQI $\uparrow$    &   0.994 &   0.993 &   0.991 &    0.991 \\
    DD $\downarrow$   &   0.018 &   0.024 &   0.027 &    0.027 \\
    CC $\uparrow$     &   0.944 &   0.931 &   0.890 &    0.884 \\
    \bottomrule
    \end{tabular}}
\end{table*}


It is worth and necessary to underline that for the found state-of-the-art works in clouds-removing the only work that released the trained weights is \cite{meraner2020cloud}, where authors released within their GitHub page, a proper guide to use their model.

Qualitative results of the proposed method and the relative comparisons with DSen2-CR from \cite{meraner2020cloud} are shown in Figure \ref{fig:comparison}, where the image \ref{fig:comparison}-(a) is the ground truth acquired with a slightly temporal delay from the cloudy image \ref{fig:comparison}-(b) in order to get a cloud-free acquisition. The PLFM prediction is shown in Figure \ref{fig:comparison}-(c), while the DSen2-CR one in Figure \ref{fig:comparison}-(d). The corresponding quantitative results are reported in Table \ref{tab:comparison}.

The results show the ability of the proposed method in reconstructing the optical images even with a large cloud coverage. Moreover, the proposed model outperforms the DSen2-CR, by showing good improvements in all the metrics.

\begin{table}[!ht]
    \centering
    \caption{Quantitative results where the proposed model is compared with DSen2-CR\cite{meraner2020cloud}}\label{tab:quantRes2}
    \resizebox{0.9\columnwidth}{!}{
    \begin{tabular}{lccc}
    \toprule
         Ground Truth vs. & Cloudy & \textbf{PLFM} & DSen2-CR\cite{meraner2020cloud} \\
    \midrule
         PSNR $\uparrow$     & 13.103 & \textbf{30.648} & 17.024 \\
         SSIM $\uparrow$    &  0.720 & \textbf{ 0.923} &  0.840 \\
         RMSE $\downarrow$    &  0.221 & \textbf{ 0.029} &  0.141 \\
         SAM  $\downarrow$    &  0.122 & \textbf{ 0.092} &  0.131 \\
         UQI  $\uparrow$    &  0.790 & \textbf{ 0.994} &  0.890 \\
         DD   $\downarrow$    &  0.196 & \textbf{ 0.021} &  0.120 \\
         CC   $\uparrow$    &  0.881 & \textbf{ 0.941} &  0.920 \\
    \bottomrule
    \end{tabular}}
    \label{tab:comparison}
\end{table}

In Table \ref{tab:pubres} we report the published results of   state of the art methods for cloud removal. Unfortunately, only the work of \cite{meraner2020cloud} releases network weights. Comparisons with published performances are not truly fair, since the models are not tested under the same settings (e.g. different training/testing images, different cloud coverage). Several metrics mitigate the problem. As anticipated in the introduction  preserving the details is a critical action and specific metrics are needed beyond the typical SSIM and PSNR. 

An example is given in  Table \ref{tab:pubres} where the performance of our method compared with others  have lower values, for some of the metrics. However, we computed the cloud coverage  of the images shown in the paper, obtaining approximate estimation of the coverage. Our evaluation of the cloud coverage shows that reported values, without a clear statement about coverage cannot effectively be taken into account.

\begin{table}[!ht]
    \centering
    \caption{Publicly available results for the state-of-the-art methods on cloud removal. When the cloud coverage percentage is not available we estimate it by counting white pixels.}\label{tab:pubres}
    \resizebox{1\columnwidth}{!}{
    \begin{tabular}{lccccc}
    \toprule
     Method & PSNR $\uparrow$ & SSIM $\uparrow$ & SAM $\downarrow$ & RMSE $\downarrow$ & Cloud Cover\\
       &   &   &   &   & percentage \\
    \midrule
   Chen et. \textit{Al} \cite{11}            & \textbf{36.14} & -    & -    & -  & $8$ \\
   Gao et. \textit{Al} \cite{gao2020cloud}   &     - & 0.97 & 3.16 & 9.79 & $19$\\
   Liu. \textit{Al} \cite{24}                & 33.25 & \textbf{0.98} & -    & 5.35 & $14$\\
   Li et. \textit{Al} \cite{25}              & 22.11 & 0.84 &    - & -  &  $23$\\
   Zheng et. \textit{Al} \cite{26}           & 20.01 & 0.81 &    - & -  &  $ 55$\\
   Zhao et. \textit{Al} \cite{27}            & 28.16 & 0.79 &    - & -  &  $ 5$\\
   Grohnfeldt et. \textit{Al} \cite{grohnfeldt2018conditional} & -     & -    & 11.0 & 0.43 & $ 16$\\
   Sarukkai et. \textit{Al} \cite{sarukkai2020cloud}  & 26.18 & 0.73 &    - & - &    $ 30$\\
   Oehmcke et. \textit{Al} \cite{oehmcke2020creating}   & -     & 0.55 & -    & - &   $35$\\
   
   PLFM & 30.65 & 0.94 & \textbf{0.09} & \textbf{0.03} & $70$ \\
   \bottomrule
   \end{tabular}}
\end{table}

\subsection{Experiments on thin and thick clouds}
In this section some experiments on thin and thick clouds were reported with the intent of showing that the proposed model is able to correctly restore the cloudy image under several cloud coverage conditions. 

\begin{figure}[!ht]
    \centering
    \resizebox{1\columnwidth}{!}{
    \begin{tabular}{ccccc}
          & \scalebox{2}{\textbf{Ground Truth}} & \scalebox{2}{\textbf{Cloudy}} & \scalebox{2}{\textbf{PLFM}} & \scalebox{2}{\textbf{DSen2-CR}\cite{meraner2020cloud}}  \\
         \scalebox{2}{\textbf{A}}     & \includegraphics[width=0.5\columnwidth]{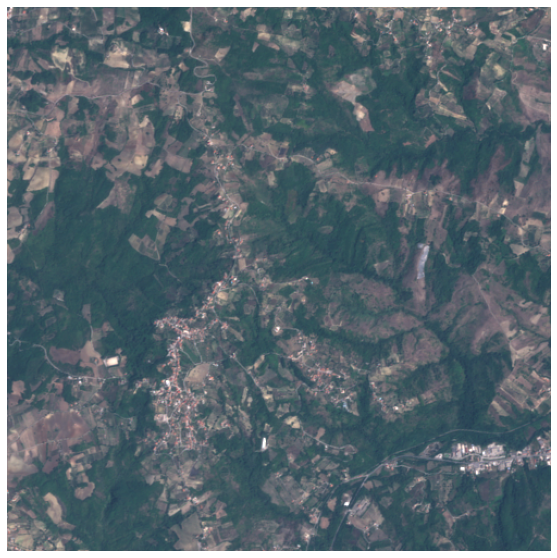}
                 & \includegraphics[width=0.5\columnwidth]{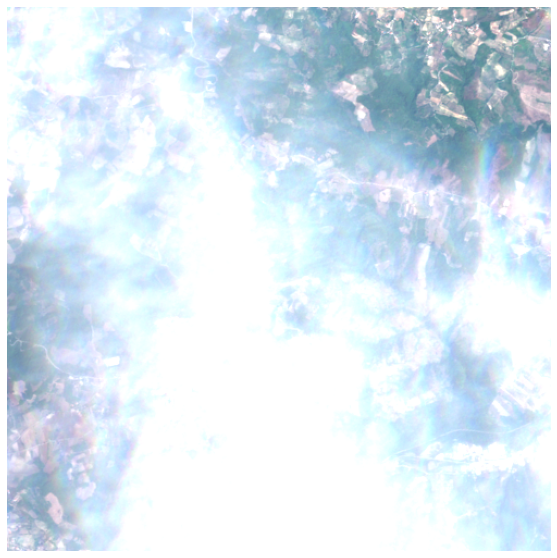}
                 & \includegraphics[width=0.5\columnwidth]{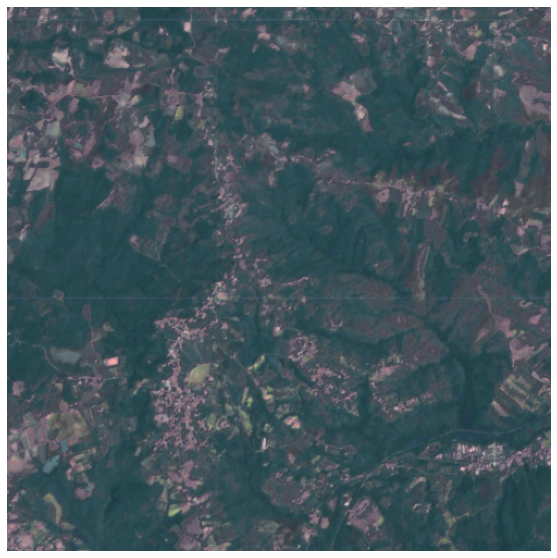}
                 & \includegraphics[width=0.5\columnwidth]{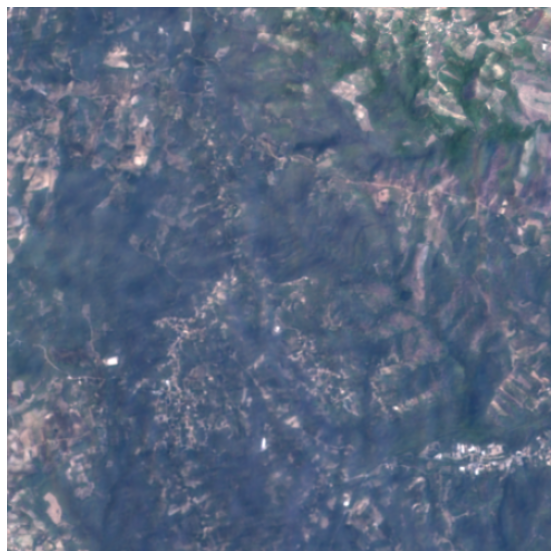}\\
         \scalebox{2}{\textbf{B}}     & \includegraphics[width=0.5\columnwidth]{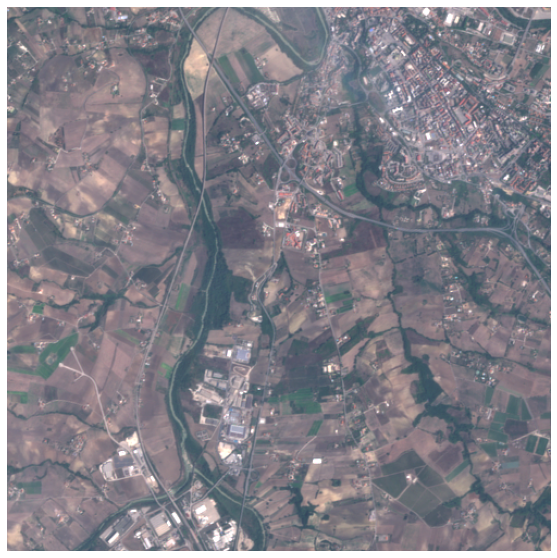}
                 & \includegraphics[width=0.5\columnwidth]{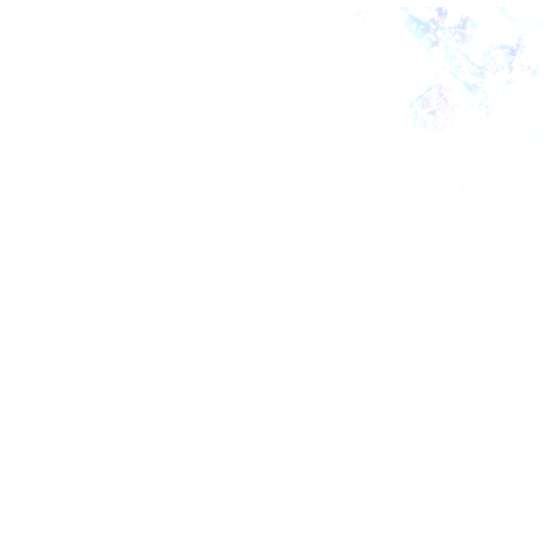}
                 & \includegraphics[width=0.5\columnwidth]{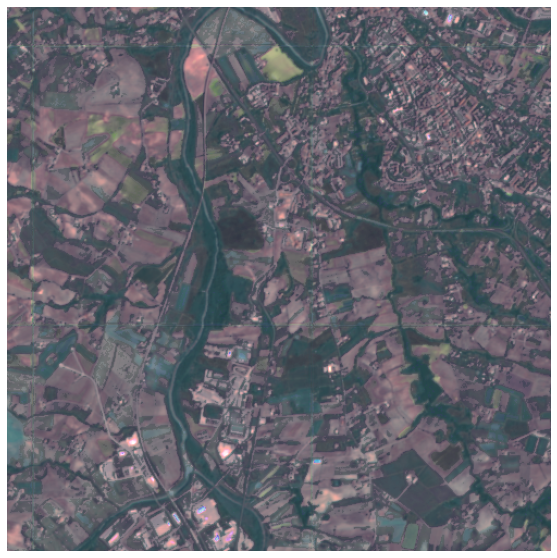}
                 & \includegraphics[width=0.5\columnwidth]{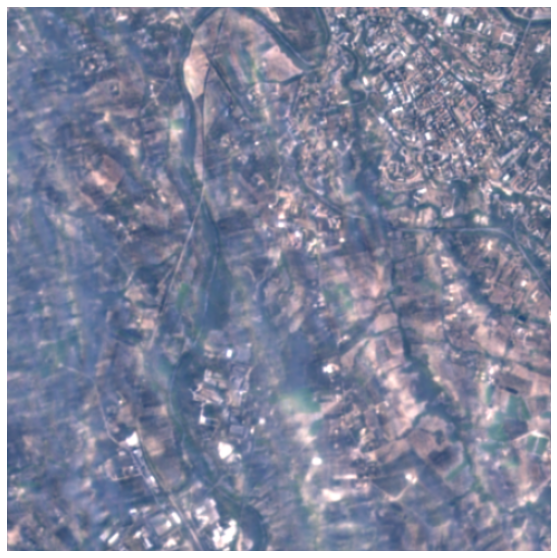}\\
         \scalebox{2}{\textbf{C}}     & \includegraphics[width=0.5\columnwidth]{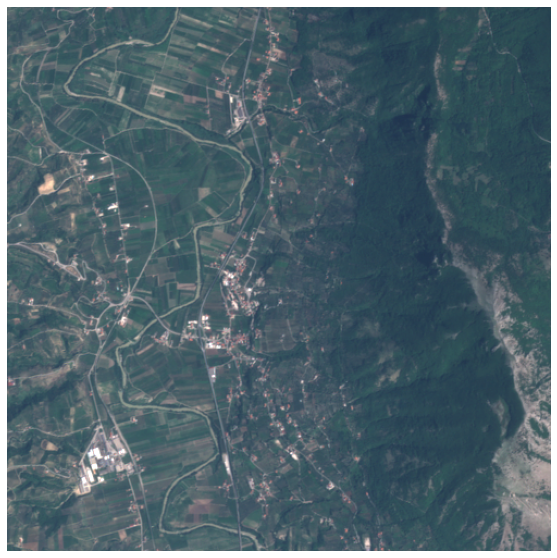}
                 & \includegraphics[width=0.5\columnwidth]{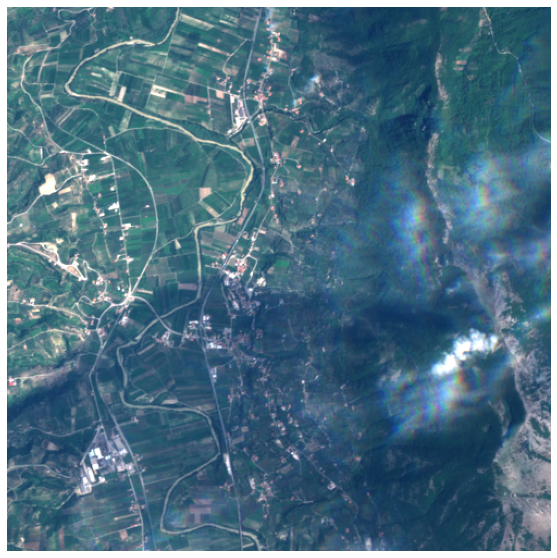}
                 & \includegraphics[width=0.5\columnwidth]{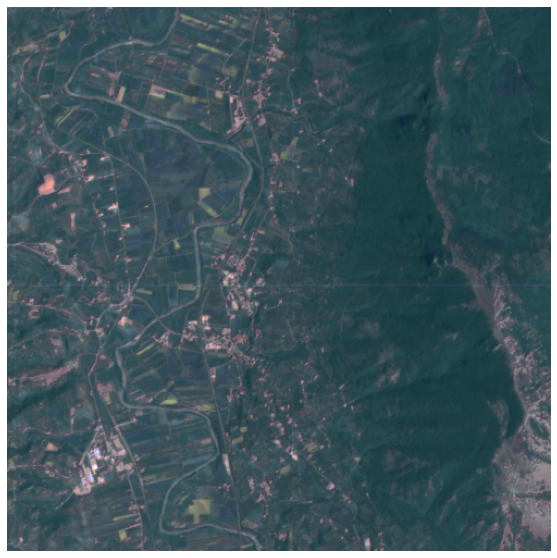}
                 & \includegraphics[width=0.5\columnwidth]{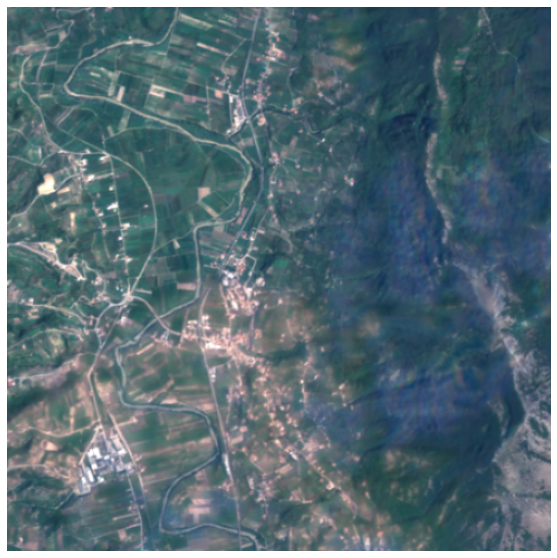}\\
         \scalebox{2}{\textbf{D}}     & \includegraphics[width=0.5\columnwidth]{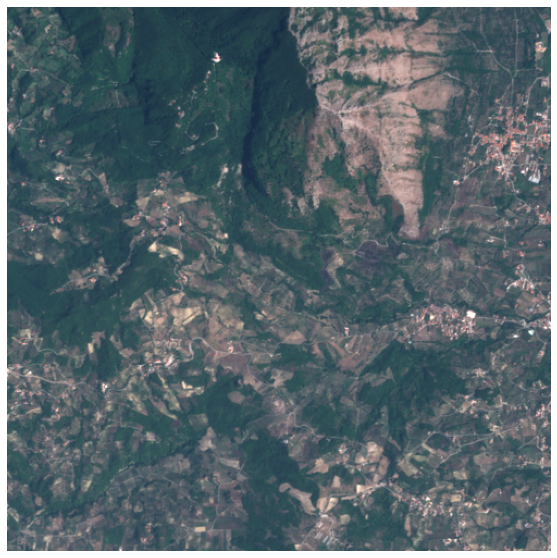}
                 & \includegraphics[width=0.5\columnwidth]{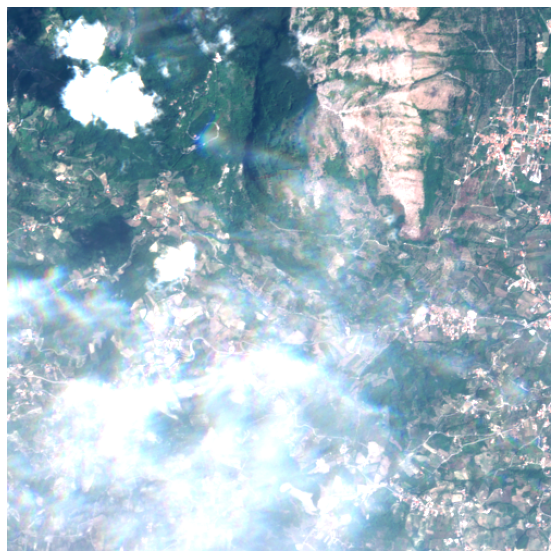}
                 & \includegraphics[width=0.5\columnwidth]{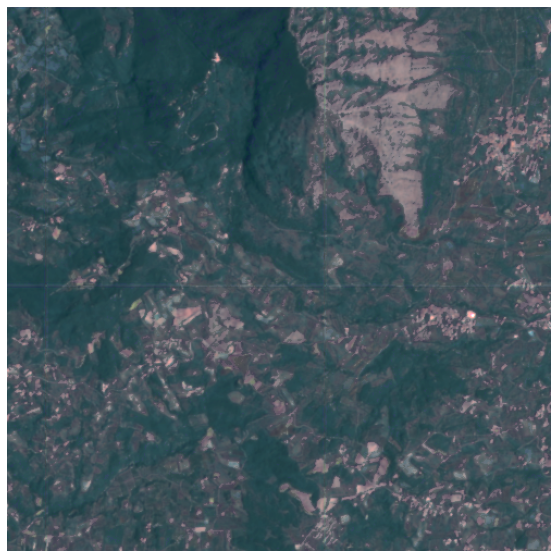}
                 & \includegraphics[width=0.5\columnwidth]{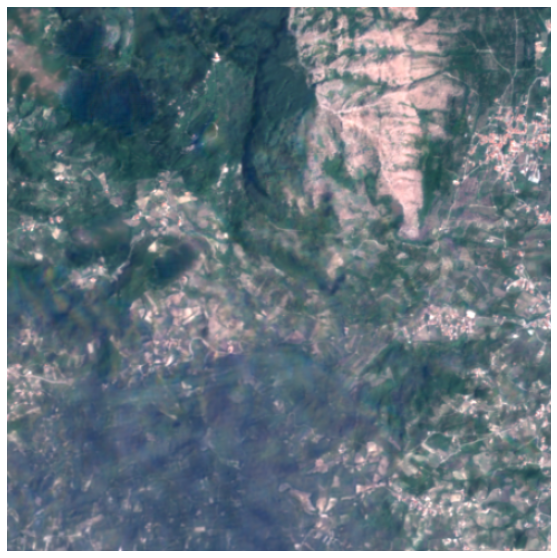}\\
    \end{tabular}}
    \caption{Qualitative experiments with thin and thick clouds on test set. Each row represent a specific experiment, while each column represents respectively the ground truth, the cloudy image, the PLFM prediction and the DSen2-CR \cite{meraner2020cloud} prediction.}
    \label{fig:thintickclouds}
\end{figure}

In Figure \ref{fig:thintickclouds} we report four experiments on thin and thick clouds, where each row represents an experiment and the columns represent respectively the ground truth, the cloudy image to be restored, the PLFM prediction and DSen2-CR prediction \cite{meraner2020cloud}. Corresponding quantitative results can be found in Table \ref{tab:thintickclouds}. Even if  DSen2-CR shows a good ability of restoring the cloudy image, our model outperform it.

\begin{table}[!ht]
    \centering
    \caption{Quantitative experiments with thin and thick clouds}
    \resizebox{0.9\columnwidth}{!}{
    \begin{tabular}{llc >{\bfseries}c c}
    \toprule
    Val & Ground Truth vs. & Cloudy & PLFM & DSen2-CR \cite{meraner2020cloud}\\
    \midrule
         & PSNR $\uparrow$    &   3.695 &  30.315 &  16.697  \\      
         & SSIM $\uparrow$    &   0.468 &   0.939 &   0.836  \\      
        & RMSE $\downarrow$   &   0.653 &   0.030 &   0.146  \\  
     A    & SAM  $\downarrow$    &   0.220 &   0.090 &   0.165  \\     
         & UQI $\uparrow$     &   0.319 &   0.992 &   0.854  \\      
         & DD   $\downarrow$    &   0.648 &   0.022 &   0.133  \\      
         & CC  $\uparrow$     &   0.357 &   0.908 &   0.706  \\      
    \midrule
         & PSNR $\uparrow$    &   4.113 &  27.373 &  16.330  \\      
         & SSIM $\uparrow$    &   0.497 &   0.915 &   0.788  \\       
        & RMSE $\downarrow$    &   0.623 &   0.043 &   0.153  \\  
    B     & SAM  $\downarrow$    &   0.180 &   0.093 &   0.161  \\      
         & UQI  $\uparrow$    &   0.441 &   0.988 &   0.904  \\       
         & DD   $\downarrow$    &   0.619 &   0.034 &   0.131  \\      
         & CC   $\uparrow$    &   0.000 &   0.897 &   0.591  \\      
    \midrule
         & PSNR  $\uparrow$   &  14.036 &  30.444 &  17.604 \\
         & SSIM  $\uparrow$   &   0.731 &   0.921 &   0.851 \\ 
         & RMSE  $\downarrow$   &   0.199 &   0.030 &   0.132 \\
    C     & SAM   $\downarrow$   &   0.127 &   0.098 &   0.116 \\
         & UQI $\uparrow$     &   0.805 &   0.993 &   0.893 \\
         & DD   $\downarrow$    &   0.175 &   0.022 &   0.115 \\
         & CC  $\uparrow$     &   0.942 &   0.859 &   0.922 \\
    \midrule
         & PSNR $\uparrow$    &   8.852 &  30.161 &  17.276 \\
         & SSIM $\uparrow$    &   0.601 &   0.946 &   0.815 \\
         & RMSE $\downarrow$    &   0.361 &   0.031 &   0.137 \\
    D     & SAM  $\downarrow$    &   0.372 &   0.100 &   0.174 \\
         & UQI $\uparrow$     &   0.563 &   0.987 &   0.835 \\
         & DD  $\downarrow$     &   0.299 &   0.025 &   0.119 \\
         & CC  $\uparrow$     &   0.408 &   0.923 &   0.853 \\
    \bottomrule
    \end{tabular}}
    
    \label{tab:thintickclouds}
\end{table}

\begin{table}[!ht]
    \centering
    \caption{Quantitative results with CSC and without CSC}\label{tab:co-not-co}
    \resizebox{1\columnwidth}{!}{
    \begin{tabular}{lccc}
    \toprule
         Ground Truth vs. & Cloudy & \textbf{PLFM} &  PLFM\\
           &  & \textbf{CSC} &  no-CSC\\
    \midrule
         PSNR $\uparrow$    & 13.103 & \textbf{30.648} & 30.547 \\
         SSIM  $\uparrow$   &  0.720 & \textbf{ 0.923} &  0.901 \\
         RMSE  $\downarrow$   &  0.221 & \textbf{ 0.029} &  0.038 \\
         SAM   $\downarrow$   &  0.122 & \textbf{ 0.092} &  0.098 \\
         UQI   $\uparrow$   &  0.790 & \textbf{ 0.994} &  0.990 \\
         DD   $\downarrow$   &  0.196 & \textbf{ 0.021} &  0.031 \\
         CC    $\uparrow$   &  0.881 & \textbf{ 0.941} &  0.889 \\
    \bottomrule
    \end{tabular}}
    \label{tab:comparison}
\end{table}

\begin{table}[!ht]
   \centering
   \caption{Quantitative results without LSTM}\label{tab:noLSTM}
   \resizebox{1\columnwidth}{!}{
   \begin{tabular}{lcccc}
   \toprule
         Ground Truth vs. & Cloudy & \textbf{PLFM} & GAN-Pix2Pix\\
    \midrule
         PSNR  $\uparrow$   & 13.103 & \textbf{30.648} & 25.331\\
         SSIM  $\uparrow$   &  0.720 & \textbf{0.923} & 0.910\\
         RMSE $\downarrow$    &  0.221 & \textbf{0.029} & 0.054\\
         SAM  $\downarrow$    &  0.122 & \textbf{0.092} & 0.093\\
         UQI  $\uparrow$    &  0.790 & \textbf{0.994} & 0.961\\
         DD   $\downarrow$    &  0.196 & \textbf{0.021} & 0.043\\
         CC   $\uparrow$    &  0.881 & \textbf{0.941} & 0.890\\
    \bottomrule
    \end{tabular}}
    
\end{table}
\begin{table}[!ht]
   \centering
   \caption{Quantitative results without Pix2Pix GAN} \label{tab:noGAN}
   \resizebox{1\columnwidth}{!}{
   \begin{tabular}{lcccc}
    \toprule
         Ground Truth vs. & Cloudy & \textbf{PLFM} & CNN-LSTM\\
    \midrule
         PSNR $\uparrow$    & 13.103 & \textbf{30.648} & {26.061} \\
         SSIM  $\uparrow$   &  0.720 & \textbf{ 0.923} & { 0.921} \\
         RMSE $\downarrow$    &  0.221 & \textbf{ 0.029} & { 0.050} \\
         SAM   $\downarrow$   &  0.122 & \textbf{ 0.092} & { 0.083} \\
         UQI   $\uparrow$   &  0.790 & \textbf{ 0.994} & { 0.971} \\
         DD   $\downarrow$    &  0.196 & \textbf{ 0.021} & { 0.040} \\
         CC    $\uparrow$   &  0.881 & \textbf{ 0.941} & { 0.911} \\
    \bottomrule
    \end{tabular}}
\end{table}
  
\begin{table}[!ht]
    \centering
    \caption{Quantitative results for SAR-OPT-LSTM}\label{tab:SAR-OPT-LSTM}
    \resizebox{1\columnwidth}{!}{
    \begin{tabular}{lccc}
    \toprule
         Ground Truth vs. & Cloudy & \textbf{PLFM} & SAR-OPT-LSTM\\
    \midrule
         PSNR  $\uparrow$   & 13.103 & \textbf{30.648} & 25.220\\
         SSIM  $\uparrow$   &  0.720 & \textbf{ 0.923} & 0.901\\
         RMSE $\downarrow$    &  0.221 & \textbf{ 0.029} & 0.061\\
         SAM  $\downarrow$    &  0.122 & \textbf{ 0.092} & 0.102\\
         UQI  $\uparrow$    &  0.790 & \textbf{ 0.994} & 0.907\\
         DD   $\downarrow$    &  0.196 & \textbf{ 0.021} & 0.042\\
         CC   $\uparrow$    &  0.881 & \textbf{ 0.941} & 0.881\\
    \bottomrule
    \end{tabular}}
\end{table}

\subsection{Ablation}
To conclude the experimental analysis a proper ablation study is described in this section.  Essentially we do three kind of ablations. The first ablation studies the difference between results when the CSC is applied or not. Results are shown in Table \ref{tab:co-not-co}, where we can see that despite there is a drop in performance this is minimal.

In the second ablation we study reduced models. Namely, we remove some component from the multimodal model and  test the remaining components  demonstrating that the complete model performs better than each reduced network. In particular we show the following ablated networks:

\begin{enumerate}
    \item removing the convLSTM. Results are reported in Table \ref{tab:noLSTM}; 
    \item removing the cGAN. Results are reported in Table \ref{tab:noGAN}.
\end{enumerate}

Tests were done using cloud coverage as the one shown in Figure \ref{fig:results}-(b) and the corresponding ground truth Figure \ref{fig:results}-(a). Comparing with Table \ref{tab:quantRes2} we can note that all the mean values  loose performance, especially for PSNR and CC.

The third ablation study is done by modifying the proposed convLSTM, which we define SAR-OPT-LSTM, that takes both SAR and Optical data as input. As expected the SAR-OPT-LSTM is not able to blend SAR information correctly and consequently it is possible to notice a drop in performance as shown in  Table \ref{tab:SAR-OPT-LSTM}.






\section{Discussion and Conclusions}\label{sec:conclusion_discussion}

 
In this manuscript we propose a novel model for cloud removal  called PLFM. The model has three branches, two branches elaborate  two diverse inputs formed by a temporal sequence of optical images  and a SAR image. The third branch generates a new optical image free of clouds. The proposed multimodal model ensures to generate cloud free images from input images covered up to 99\%.
Each branch of the network is a deep network with its own loss function and minimization to cope with the idiosyncratic properties of the inputs.  We treat the first two branches as building an embedding into smoothed cloudy images, namely as intermediate feature maps for the multimodal model head that generates the cloud free image. The multimodal model has a unique flow from the input up to the cloud free generated image.
 
 The model uses as baselines for the branches out off-the-shelf   networks such as convLSTM, conditional GANs and encoder-decoder CNN. 

Our results show that the proposed method is competitive with the current state of the art producing  realistic results in several cloud coverage conditions, which we prove both with quantitative tables and qualitative images.  An ablation study shows that  removing any part of the architecture obtains models lowering the performance of PFLM, we have also  shown that a  ConvLSTM when both SAR and optical data are considered as input, cannot obtain the same performance as PFLM..

Besides these novelties, we have also built a dataset pairing S1 and S2 images randomly distributed on the whole Earth surface, in order to promote the model generalization. The dataset is publicly available.

Future developments of our work will include exploring more complex datasets (e.g. multispectral, multipolarimetric, and longer time-series), different parameters and structure with the objective to further optimize the proposed method, and exploiting the SAR temporal evolution.

\section*{Acknowledgments}
The research work published in this manuscript has been developed in collaboration with the European Space Agency (ESA) $\Phi$-lab during the traineeship of Maria Pia Del Rosso and Alessandro Sebastianelli in the period July-October 2019 and during the traineeship of Erika Puglisi in the period February-October 2020. This research is also supported by the ongoing Open Space Innovation Platform (OSIP) project started in June 2020 and entitled "Al powered cross-modal adaptation techniques applied to Sentinel-1 and -2 data",  under a joint collaboration between the ESA $\Phi$-Lab and the University of Sannio. Part of this work has been awarded with one of the IEEE GRS29-Italy 2020 prizes.

\appendices

\section{Math preliminary on LSTM and cGAN}\label{sec:appendix_math_preliminary_GAN_LTM}
In this section we provide some details on the deep learning baseline models used in the proposed model for cloud removal. In particular, we recap basic ideas of GANs \cite{goodfellow2014generative}, cGANs \cite{isola2017image,zhu2017unpaired,lin2018conditional} and  ConvLSTM \cite{hochreiter1997long,xingjian2015convolutional}.

\subsection{Generative Adversarial Networks}
Generative adversarial networks (GANs) are deep learning models, which have shown to be very powerful to learn distributions for synthesizing new images \cite{goodfellow2014generative}. GANs, based on a generator taking random noise as input and on a discriminator forcing  the generator to refine the output distribution via an {\it adversarial loss}, are able to generate perceptually realistic new images. Relevant and very successful GAN-based methods developed the idea of training the GANs to translate images between two different domains. 

Despite the stringent requirements needed to perfectly match images from two different domains \cite{isola2017image}, the cross-domain translation has been successfully addressed initially with a Pix2Pix model, and then with the CycleGAN \cite{zhu2017unpaired}, DualGAN \cite{yi2017dualgan} and CoGAN \cite{liu2016coupled} model, able to accept unpaired images.

As stated before a GAN combines discriminative and generative phenomena. The discriminative component aims to learn the conditional probability distribution $p(y|x)$, so the model learns to predict the output $y$ given the input data $x$. On the other hand, the generative components aims to learn the joint probability distribution $P(x,y)=p(x|y)p(y)$, so the model is trained to understand the inputs to generate similar inputs and it’s labels.

Let $z$ be a noise vector, $G(z)=x_{fake}$ be a generator's output, $x=x_{real}$ be a training sample, $D(x) \rightarrow p(y|x_{real})$ be the discriminator output for $x_{real}$ and $D(G(z)) \rightarrow p(y|x_{fake})$. The goal is to build a discriminator, that maximizes the real data while minimizing the fake data, and a generator, that maximized the fake data. In other words, for the discriminator $D$, $D(x)$ should be maximized and $D(G(z))$ should be minimized and for the generator $G$, $D(G(z))$ should be maximized

The original GAN \cite{goodfellow2014generative} presents a discriminator loss given by:
\begin{equation}
    \centering
    \begin{split}
    D_{loss} &= D_{loss_{real}} + D_{loss_{fake}}=\\
    &= \log(D(x)) + \log(1-D(G(z)))\\    
    \end{split}
\end{equation}

generalizable to the case with n samples

\begin{equation}
    \centering
    D_{loss}=\frac{1}{n}\sum_{i=1}^{n}{\log(D(x^i))+\log(1-D(G(z^i)))}
\end{equation}

and a generator loss given by

\begin{equation}
    \centering
    G_{loss} = \log(1-D(G(z))) = -\log(D(G(z)))
\end{equation}

generalizable to the case with n samples

\begin{equation}
    \centering
    \begin{split}
    G_{loss} &= \frac{1}{n}\sum_{i=1}^{n}{\log(1-D(G(z^i)))} =\\
    &=\frac{1}{n}\sum_{i=1}^{n}{-\log(D(G(z^i)))}\\
    \end{split}
\end{equation}

Given the above equation the min-max problem involving the discriminator and the generator can be expressed with

\begin{equation}
    \centering
    \begin{split}
    &\min_{G}\max_{D}V(D,G) =\\
    &=\mathbb{E}_{x\sim p_{x}(x)}[\log (D(x))] + \mathbb{E}_{z\sim p_{z}(z)}[\log (1- D(G(z)))]
    \end{split}
\end{equation}

or optionally

\begin{equation}
    \centering
    \begin{split}
    &\max_{D}V(D)=\\
    &=\underbrace{\mathbb{E}_{x\sim p_{x}(x)}[\log (D(x))]}_{\text{recognize real images better}} + \underbrace{\mathbb{E}_{z\sim p_{z}(z)}[\log (1- D(G(z)))]}_{\text{recognize generated images better}}\\
    &\min_{G}V(G)=\underbrace{\mathbb{E}_{z\sim p_{z}(z)}[\log (1- D(G(z)))]}_{\text{optimize G to fool D the most}}\\
    \end{split}
\end{equation}

In the cGANs the adversarial loss is reformulated to take into account the principle of conditioning on the observed images. In the generator for cGANs, the prior input noise $p_z(z)$ and the auxiliary information $y$ are combined in joint hidden representations. In the discriminator, $x$ and $y$ are presented as inputs to the discriminative function. The objective function is similar to the original GAN except for the data distributions that in this case are conditioned on $y$. The cGANs objective is given by

\begin{equation}
    \centering
    \begin{split}
    &\min_{G}\max_{D}V(D,G) =\\
    &=\mathbb{E}_{x\sim p_{x}(x)}[\log (D(x|y))] + \mathbb{E}_{z\sim p_{z}(z)}[\log (1- D(G(z|y)))]
    \end{split}
\end{equation}

In an image-to-image translation the shape of the generator has a typical U-Net shape \cite{long2015fully} with skip connections to ensure information passes through all the layers including the bottleneck. In an unpaired image-to-image translation \cite{zhu2017unpaired}, together with a mapping from the observed image to the generated one, an inverse mapping from the generated image to the observed one is added. The adversarial loss is applied to both mappings in this case.

It is worth to consider that the adversarial loss has been at the heart of several variations with respect to the original GAN model, by giving rise to an incredible proliferation of different GANs.

\subsection{Convolutional Long Short-Term Memory (ConvLSTM)}
\begin{figure}[!ht]
    \centering
    \includegraphics[width=1\columnwidth]{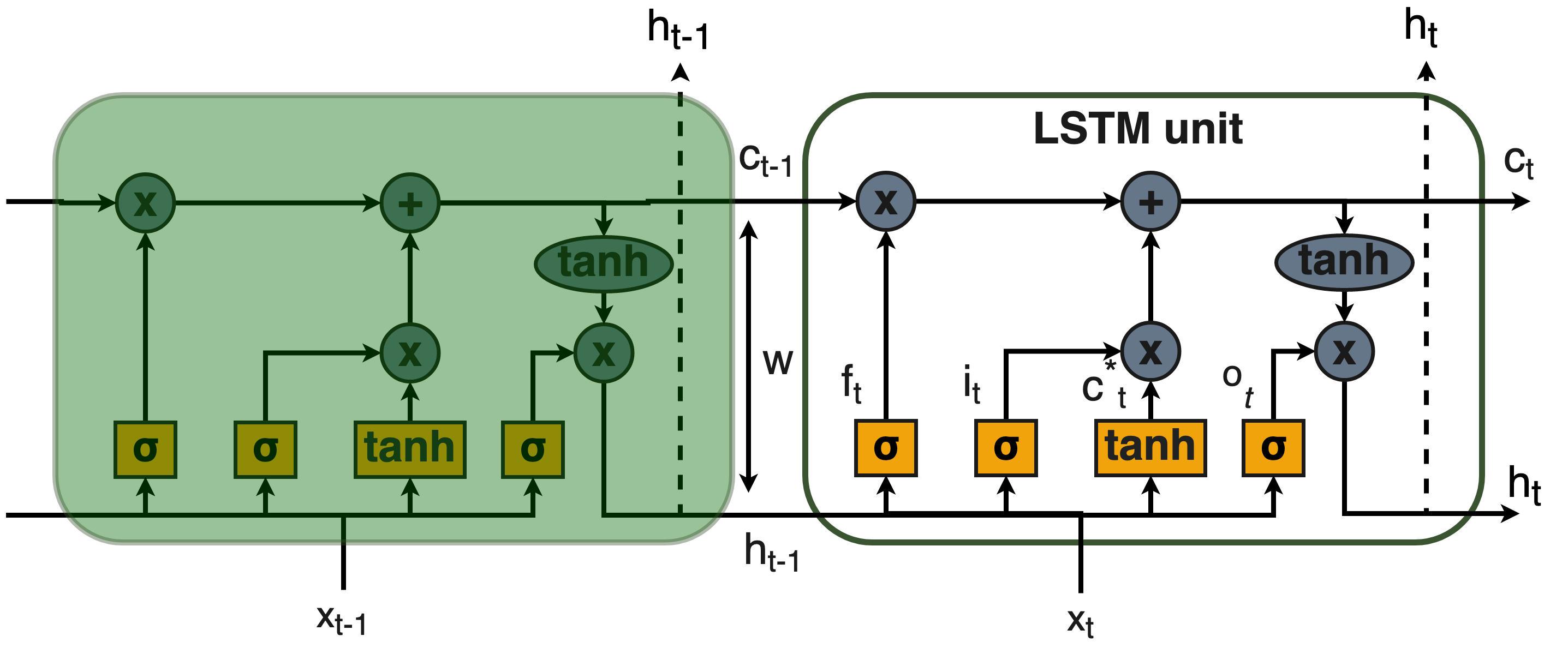}
    \caption{Classical representation of two LSTM cells. The scheme shows how the temporal input is used by the LSTM and how two adjacent cells communicates.}
    \label{fig:LSTM cell}
\end{figure}

Data acquired and collected over successive time intervals,   categorized as time-series, are typically treated with the well known Recurrent Neural Networks (RNNs) in the Machine Learning (ML) context. A particular and emerging type of RNN is the LSTM Neural Network \cite{sherstinsky2020fundamentals}. With respect to RNNs, LSTMs use an extra piece of information, called memory, for each time step in every LSTM cell. The LSTMs are formed by six main blocks: $1)$ forget gate $f$, $2)$ candidate layer $c$, $3)$ input gate $i$, $4)$ output gate $o$, $5)$ hidden state $h$ and $6)$ memory state $c$, as shown in Figure \ref{fig:LSTM cell} \cite{yu2019review, sherstinsky2020fundamentals}. 

The mathematics behind a LSTM can be summarized by the equations \eqref{eqn:lstm}, from which it is evident how this network treats temporal data:

\begin{subequations}
    \centering
    \begin{align}
        f_t &= \sigma(x_t \circ W_{xf} + h_{t-1}\circ W_{hf} + b_f)\\
        \hat{C}_t &= tanh(x_t \circ W_{xc} + h_{t-1}\circ W_{hc} + b_c)\\
        i_t &= \sigma(x_t \circ W_{xi} + h_{t-1}\circ W_{hi} + b_i)\\
        o_t &= \sigma(x_t \circ W_{xo} + h_{t-1}\circ W_{ho} + b_o)\\
        C_t &= f_t \circ C_{t-1}+ i_t \circ \hat{C}_t\\
        h_t &= o_t \circ tanh(C_t)
    \end{align}
    \label{eqn:lstm}
\end{subequations}

where $\circ$ denotes the Hadamard product, $\sigma$ denotes the sigmoid function $\sigma(x)=\frac{1}{1+e^{-x}}$, $x_t$ is the input vector, $h_{t-1}$ is the previous cell output, $C_{t-1}$ is the previous cell memory, $h_t$ is the current cell output, $C_t$ is the current cell memory, $b_{f/c/i/o}$ are bias coefficients , $W$ are the weight vectors for the forget gate, candidate gate, i/p gate, o/p gate and $\omega$ blocks represent the neural network layers.

In the Figure  \ref{fig:LSTM cell} the classical  representation  of  two  LSTM cells is shown.  The scheme  highlights   how  the  temporal  input  is  used  by  the  LSTM and how two adjacent cells communicates.

\begin{figure}[!ht]
    \centering
    \includegraphics[width=1\columnwidth]{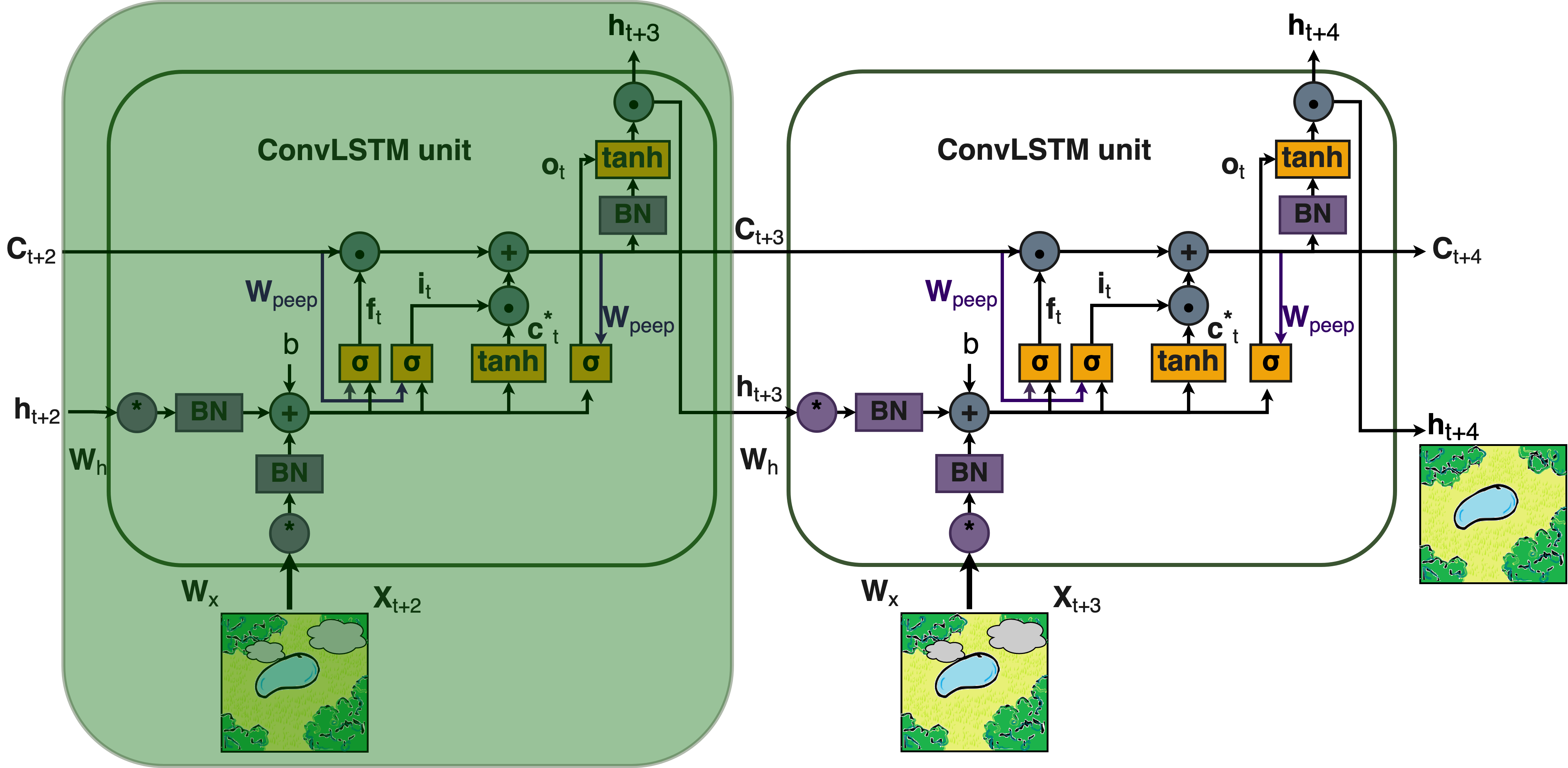}
    \caption{Classical representation of two ConvLSTM cells.  The focus is on the input, that in this case is an image, and this is reflected in the type of gates  adapted to matrix calculations.}
     \label{fig:Conv LSTM cell}
\end{figure}

The natural evolution of a LSTM is the ConvLSTM,  adopted because CNNs are the best approach when working with images. In this case the time-series are not formed anymore by vectors, but composed of images in sequence. In Figure \ref{fig:Conv LSTM cell} the typical representation of a  ConvLSTM cell is shown, where  the recurrent layer is similar to the LSTM of Figure \ref{fig:LSTM cell}, but here the focus is on the input, that in this case is an image. This is reflected in the type of gates, which are adapted to matrix calculations \cite{xingjian2015convolutional}.  

The mathematics behind a Convolutional LSTM can be summarized by:

\begin{subequations}
    \centering
    \begin{align}
    f_t   &= \sigma(W_{xf} * X_t + W_{hf} * h_{t-1} + W_{peep}\circ C_{t-1} +b_f) \\
    C^*_t &= W_{xc} * h_{t-1} + W_{hc} * h_{t-1} + b_c\\
    i_t   &= \sigma(W_{xi} * X_t + W_{hi} * h_{t-1} + W_{peep}\circ C_{t-1} + b_i)\\
    o_t   &= \sigma(W_{xo}*X_t + W_{ho} * h_{t-1} + W_{peep}\circ C_t + b_o)\\
    C_t   &= f_t*C_{t-1} + i_t \circ tanh(C^*_t)\\
    h_t   &= o_t \circ tanh(C_t)
    \end{align}
\end{subequations}

where $*$ denotes the convolution operation.
\color{black}
\bibliographystyle{IEEEtran}
\bibliography{main}
\vspace{-1cm}
\begin{IEEEbiography}
[{\includegraphics[width=1in,height=1.25in,clip,keepaspectratio]{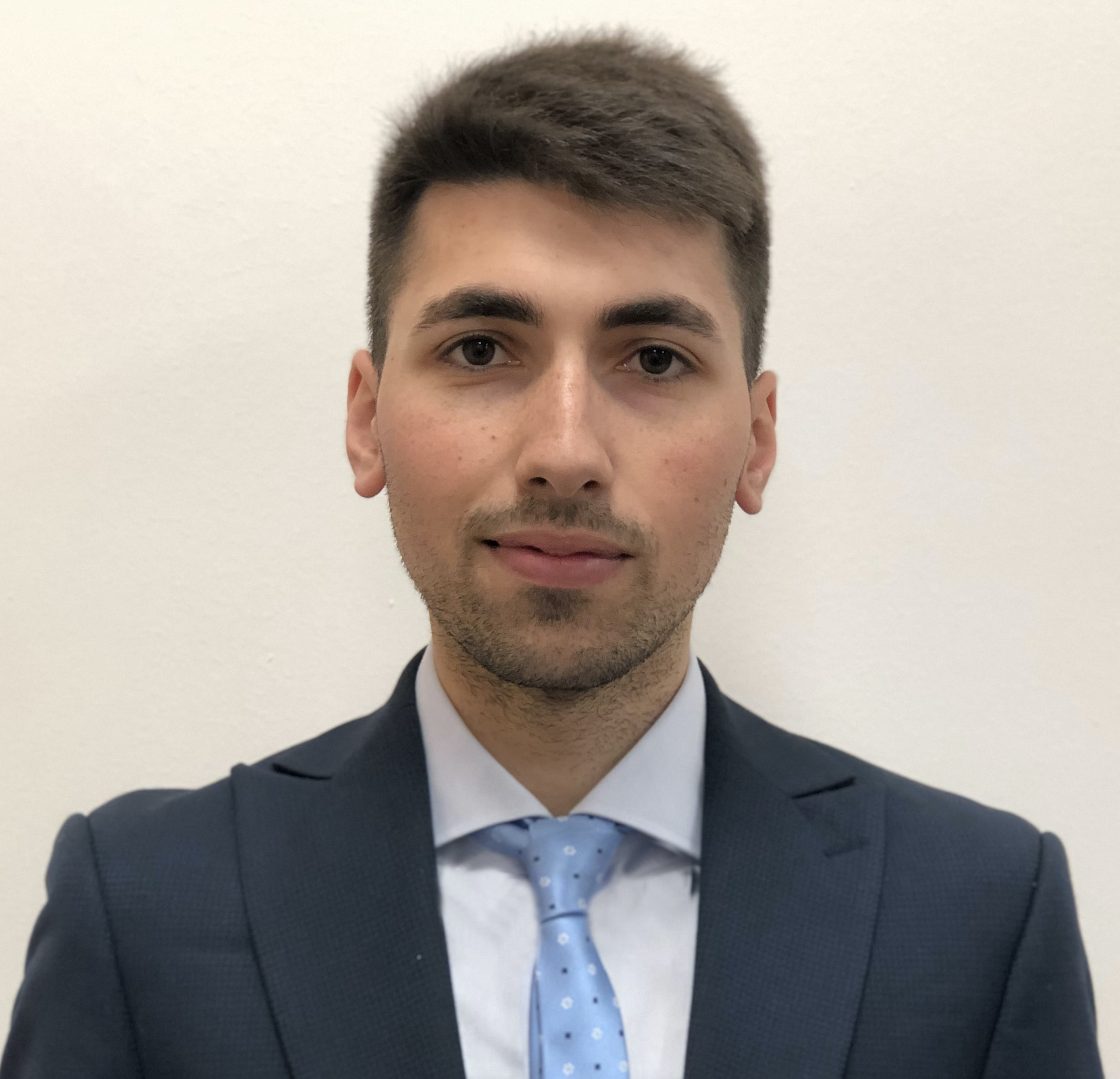}}]
{Alessandro Sebastianelli} graduated  with laude in Electronic Engineering for Automation and Telecommunications at the University  of Sannio in 2019. He is enrolled in the Ph.D. program with University of Sannio, and his research topics mainly focus on Remote Sensing and Satellite data analysis, Artificial  Intelligence  techniques for Earth Observation, and data fusion. He has co-authored several papers to reputed journals and conferences for  the  sector  of  Remote Sensing. Ha has been a visited researcher at Phi-lab in European Space  Research  Institute  (ESRIN)  of  the  European  Space Agency (ESA), in Frascati, and still collaborates with the $\Phi$-lab  on  topics  related  to  deep  learning  applied  to  geohazard assessment,  especially  for  landslides,  volcanoes,  earthquakes phenomena. He has won an ESA OSIP proposal in August 2020 presented with his Ph.D. Supervisor, Prof. Silvia L. Ullo.
\end{IEEEbiography}

\begin{IEEEbiography}[{
\includegraphics[width=1in,height=1.25in,clip,keepaspectratio]{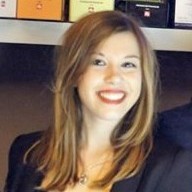}}]
{Erika Puglisi} graduated in Artificial Intelligence and Robotics at La Sapienza University of Rome in 2021. During the last year of her Master's degree, she joined as external researcher the Phi-Lab in the European Space Research Institute (ESRIN) of the European Space Agency (ESA) in Frascati. In the course of this collaboration, the topics of her research included the study of data fusion methods between SAR and optical images using Generative Adversarial Networks (GAN) and the development of deep learning algorithms for satellite data analysis. She is currently working as data engineer in the field of Business Intelligence.
\end{IEEEbiography}

\begin{IEEEbiography}[{\includegraphics[width=1in,height=1.25in,clip,keepaspectratio]{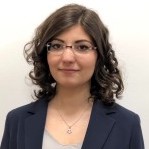}}]{Maria Pia Del Rosso} graduated  with  laude  in  Electronics Engineering for Automation and Telecommunications  at  the  University  of Sannio  in  October 2019 and she is currently a PhD candidate.  As a master student, she has been a visiting researcher at the Phi-lab in the European Space  Research  Institute  (ESRIN)  of  the  European  Space Agency (ESA), in Frascati. She worked on applying Deep Learning techniques to Remote Sensing Earth Observation data for monitoring geohazard phenomena, such as earthquakes, landslides and volcanic eruptions. She is currently working on the matching between multispectral and radar data applying AI techniques.
\end{IEEEbiography}

\begin{IEEEbiography}[{\includegraphics[width=1in,height=1.25in,clip,keepaspectratio]{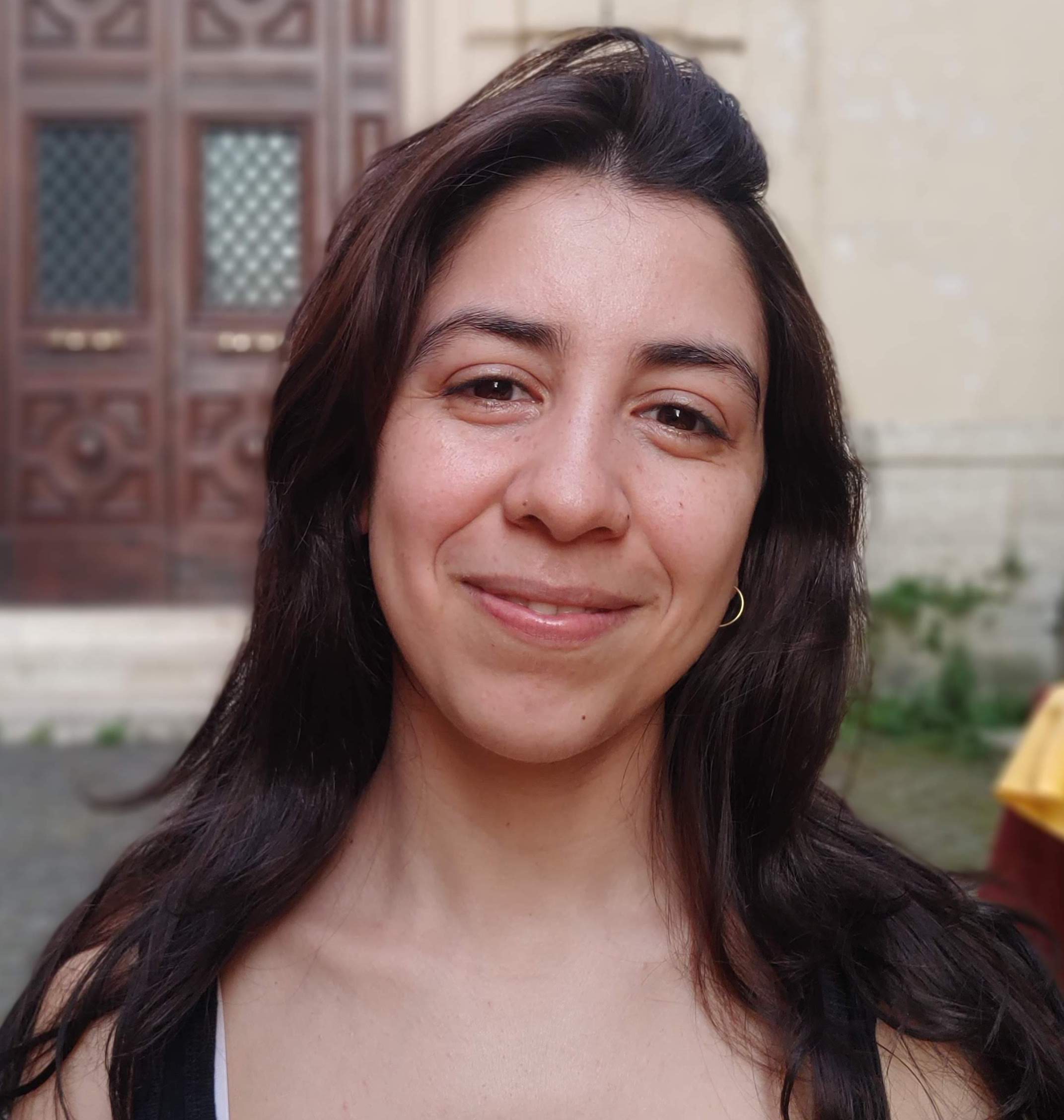}}]{Jamila Mifdal} Received an engineering diploma in electronics and computer science from ENSSAT, Lannion, France and M.Sc. degree (SISEA) in electronics and signal processing from Université Rennes 1, Rennes, France where she majored. After that she received a Ph.D. degree in applied mathemaics for satellite image processing where she developped mathematical models for data fusion. She joind the $\Phi$-lab at European Space Agency as a  researcher in artificial intelligence for Earth observation. Her research interests include image segmentation, mathematical modelling, image fusion and deep learning. The applications of her research is centered on environment monitoring.
\end{IEEEbiography}

\begin{IEEEbiography}[{\includegraphics[width=1in,height=1.25in,clip,keepaspectratio]{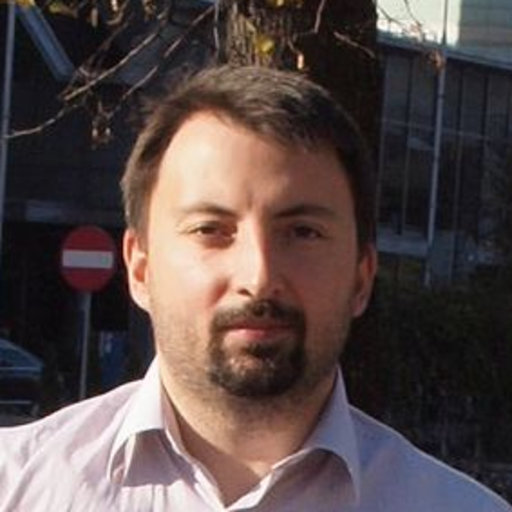}}]{Artur Nowakowski} has a background in computer science with an MSc and Ph.D. at Warsaw University of Technology.  He developed his interests in EO optical data since he joined the Space Research Centre of the Polish Academy of Science in 2011. Between 2018 and 2021 he was working as a Research Fellow for European Space Agency in the $\Phi$-lab, where he implemented new AI techniques to various kinds of EO data. Now he works for Warsaw University of Technology as a teacher and researcher and serves UN World Food Programme as a consultant in Climate and Earth Observation Team. His research include spatial analysis, time-series analysis and data fusion in the context of Earth Observations.  
\end{IEEEbiography}
\begin{IEEEbiography}[{\includegraphics[width=1in,height=1.25in,clip,keepaspectratio]{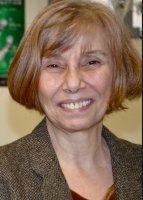}}]{Fiora Pirri}  (member IEEE) received the PhD from the UPMC (Université Paris VI Pierre et Marie Curie) now Sorbonne Université, and Master in Intelligence artificielle et reconnaissance des formes  from ENPC, Paris. She is professor at University of Sapienza, Rome. She has been the head of the Alcor Lab of vision, learning and cognitive robotics since 1998, when she founded the Lab. She has been principal investigator of several EU funded projects, in Esprit, FP5, FP6, FP7 and Horizon2020 EU programs. The most relevant projects for social impact have been NIFTI, TRADR and SecondHands. She has been awarded the bronze medal at the 2004 international world cup rescue competition.  She has been one of the winner of the Cosmo-SkyMed Italian Space Agency (ASI) International competition in 2009 with the project SARFIRE (Spaceborn SAR imagery and environmental data fusion), funded up to 2012. She has been awarded the first prize at International Symposium on Visual Computing (ISVC-2010).  With NIFTi (EU-FP6)  project she participated with Alcor Lab robots in search and rescue operations with  the Corpo Nazionale dei Vigili del fuoco (National Fire Corps), in the 2012 Mirandola earthquake. With the TRADR (EU-FP7) project Alcor Lab robots and drones participated in 2016 to the rescue operations in the Amatrice earthquacke, jointly with  the National Fire Corps.  SecondHand project (EU-H2020) has been selected as a success story for ERF 2021. She is member of the international steering committee
of Cognitive Robotics, and member of both ACM and AAAI.
\end{IEEEbiography}
\vspace{-1.8cm}
\begin{IEEEbiography}[{\includegraphics[width=1.1in,height=1.35in,clip,keepaspectratio]{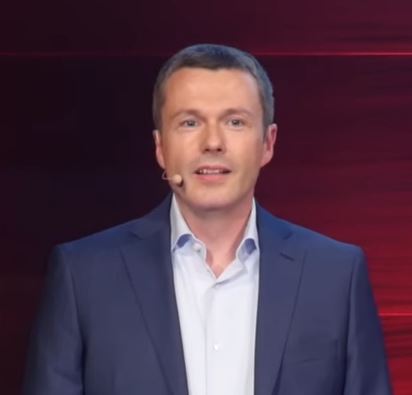}}]{Pierre Philippe Mathieu} received the M.Sc. degree in engineering from the University of Liege, Liege, Belgium, the Ph.D. degree in climate science from the University of Louvain, Louvain, Belgium, and the Management degree from the University of Reading Business School, Reading, U.K.,Over the last 20 years, he has been working in environmental monitoring and modeling, across disciplines from remote sensing, modeling, up to weather risk management. He is currently a Data Scientist with the European Space Research Institute, European Space Agency, Frascati, Italy, working to foster the use of our Earth observation missions to support science, innovation, and development in partnership with user communities, industry, and businesses. His particular interest lies in addressing global environmental and development issues related to management of foodwaterenergy resources and climate change. He is currently the head of $\Phi$-lab explore office of the European Space Agency.
\end{IEEEbiography}
\begin{IEEEbiography}[{\includegraphics[width=1in,height=1.15in,clip,keepaspectratio]{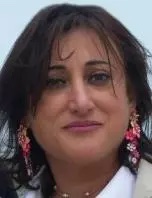}}]{Silvia Liberata Ullo} IEEE Senior Member, Industry Liaison for IEEE Joint ComSoc/VTS Italy Chapter. National Referent for FIDAPA BPW Italy Science and Technology Task Force. Researcher since 2004 in the Engineering Department of the University of Sannio, Benevento (Italy). Member of the Academic Senate and the PhD Professors’ Board. She is teaching: Signal theory and elaboration, and Telecommunication networks for Electronic Engineering, and Optical and radar remote sensing for the Ph.D. course. Authored 80+ research papers, co-authored many book chapters and served as editor of two books, and many special issues in reputed journals of her research sectors. Main interests: signal processing, remote sensing, satellite data analysis, machine learning and quantum ML, radar systems, sensor networks, and smart grids. Graduated with Laude in 1989 in Electronic Engineering,  at the Faculty of Engineering at the Federico  II  University, in  Naples, she pursued     the     M.Sc.     degree     from     the Massachusetts Institute  of Technology (MIT) Sloan  Business  School  of  Boston,  USA,  in June 1992.  She has worked in the private and public sector from 1992 to 2004, before joining the University of Sannio.
\end{IEEEbiography}

\end{document}